\documentclass[10pt,journal,compsoc]{IEEEtran}
\ifCLASSOPTIONcompsoc
  \usepackage[nocompress]{cite}
\else
  \usepackage{cite}
\fi

\usepackage{graphicx}
\usepackage{amsmath}
\usepackage{amssymb}
\usepackage{multirow}
\usepackage{caption2}
\usepackage{array}

\begin{document}

\title{Blind Image Super-Resolution: \\A Survey and Beyond}

 \author{Anran~Liu,
         Yihao~Liu,
         Jinjin~Gu,
         Yu~Qiao,
         and~Chao~Dong% <-this % stops a space
 \IEEEcompsocitemizethanks{\IEEEcompsocthanksitem A. Liu is with The University of Hong Kong, Hong Kong SAR, China.\protect\\
 E-mail: liuar616@connect.hku.hk
 \IEEEcompsocthanksitem J. Gu is with the School of Electrical and Information Engineering, The University of Sydney.\protect\\
 E-mail: jinjin.gu@sydney.edu.au
 \IEEEcompsocthanksitem Y. Liu, Y. Qiao and C. Dong are with Shenzhen Institute of Advanced Technology, Chinese Academy of Sciences, China.\protect\\
 E-mail: liuyihao14@mails.ucas.ac.cn, \{yu.qiao, chao.dong\}@siat.ac.cn}% <-this % stops a space
% \thanks{Manuscript received April 19, 2005; revised August 26, 2015.}
 }


\markboth{Journal of \LaTeX\ Class Files,~Vol.~14, No.~8, August~2015}%
{Shell \MakeLowercase{\textit{et al.}}: Bare Advanced Demo of IEEEtran.cls for IEEE Computer Society Journals}

\IEEEtitleabstractindextext{%
\begin{abstract}
Blind image super-resolution (SR), aiming to super-resolve low-resolution images with unknown degradation, has attracted increasing attention due to its significance in promoting real-world applications. Many novel and effective solutions have been proposed recently, especially with the powerful deep learning techniques. Despite years of efforts, it still remains as a challenging research problem. This paper serves as a systematic review on recent progress in blind image SR, and proposes a taxonomy to categorize existing methods into three different classes according to their ways of degradation modelling and the data used for solving the SR model. This taxonomy helps summarize and distinguish among existing methods. We hope to provide insights into current research states, as well as to reveal novel research directions worth exploring.
In addition, we make a summary on commonly used datasets and previous competitions related to blind image SR. Last but not least, a comparison among different methods is provided with detailed analysis on their merits and demerits using both synthetic and real testing images.
\end{abstract}

}

\maketitle

% is not selected <OR> if conference mode is selected - because compsoc
% conference papers position the abstract like regular (non-compsoc)
\IEEEdisplaynontitleabstractindextext

% \fi
\IEEEpeerreviewmaketitle

\ifCLASSOPTIONcompsoc
\IEEEraisesectionheading{\section{Introduction}\label{sec:introduction}}
\else
\section{Introduction}
\label{sec:introduction}
\fi

\begin{figure*}[htbp]
\renewcommand{\captionfont}{\small \rmfamily}
\begin{center}
\includegraphics[width=0.48\linewidth]{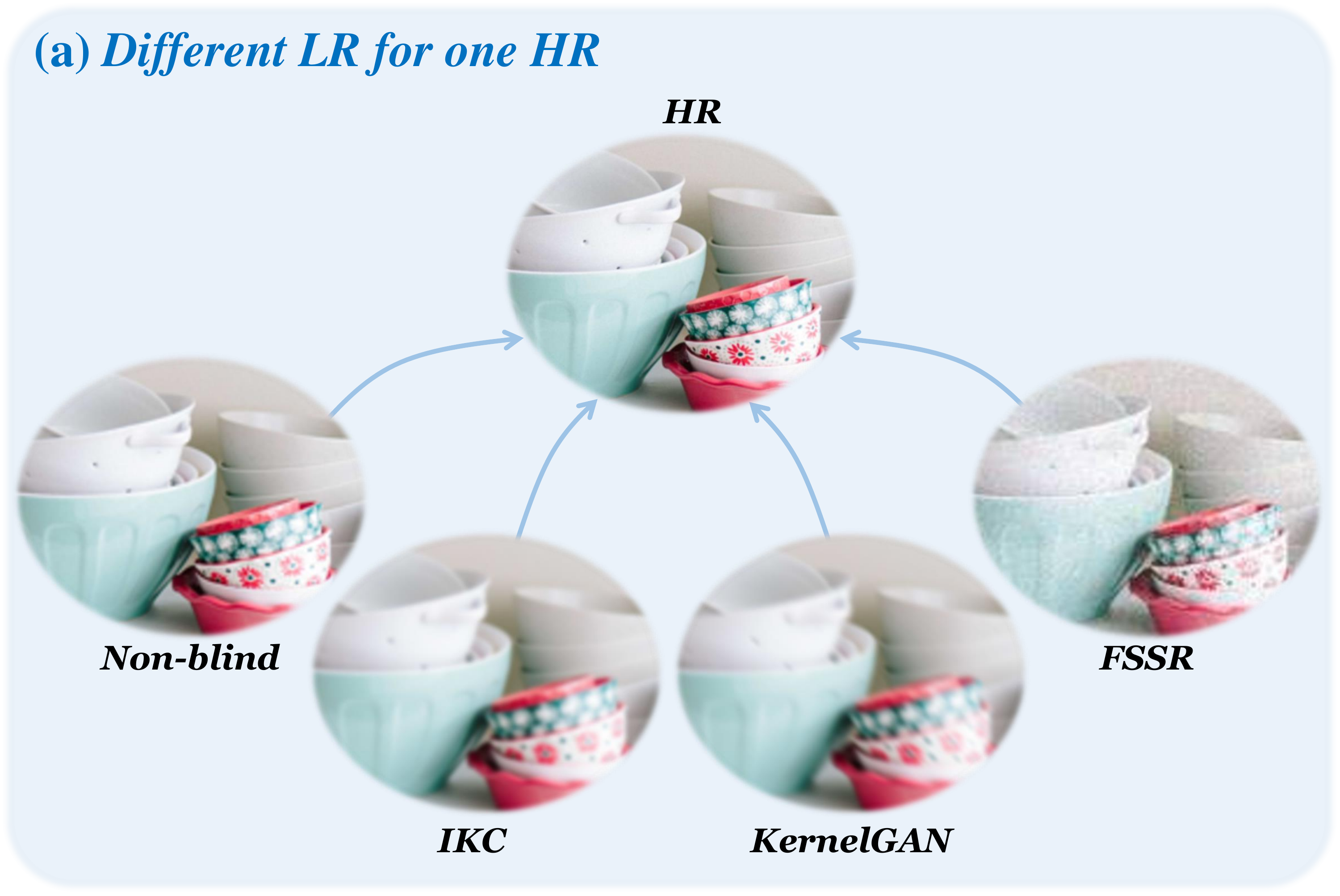}
\includegraphics[width=0.51\linewidth]{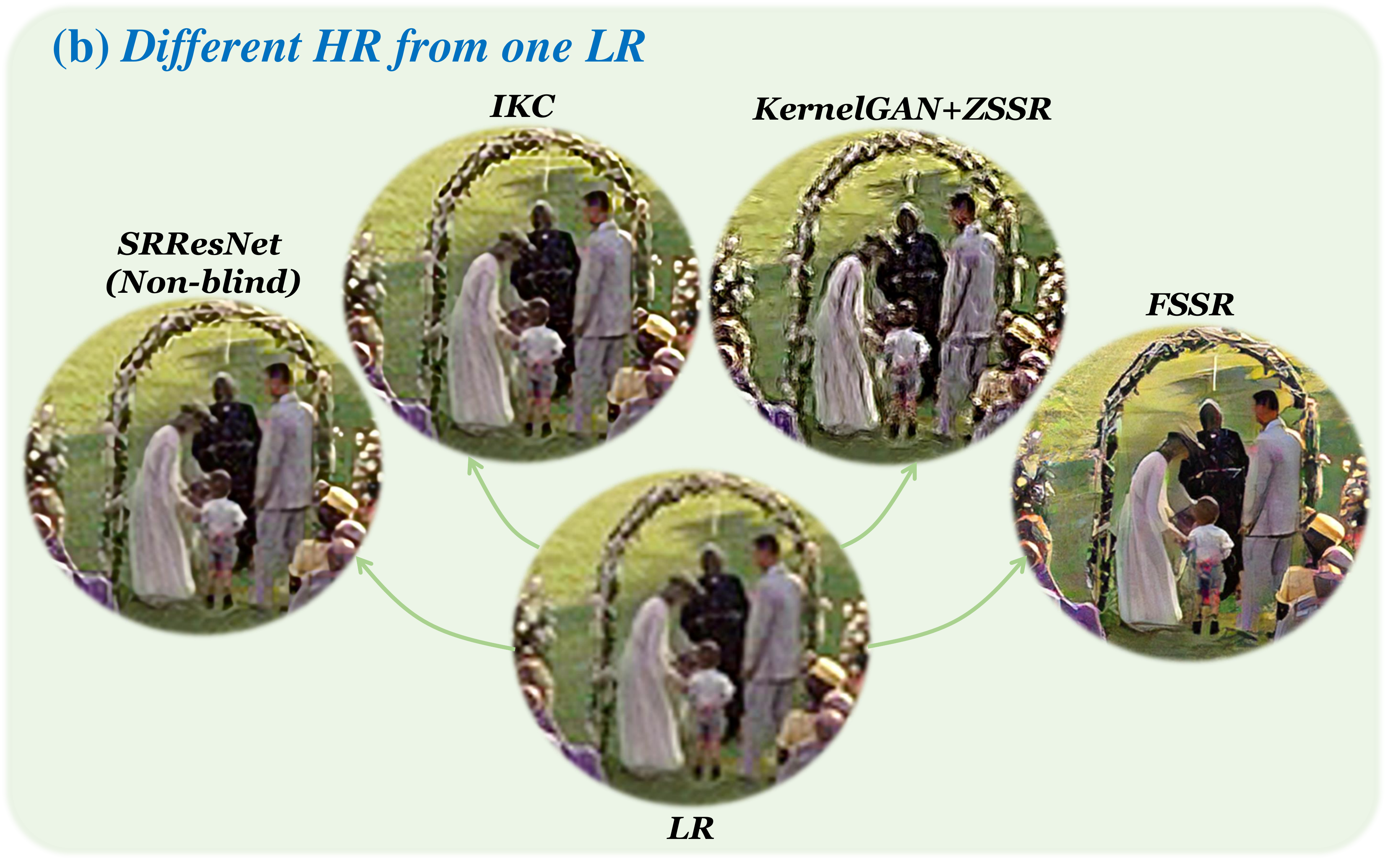}
\end{center}
\vspace{-10pt}
\caption{Left: an HR image and its different LR versions with assumed degradation types of four SISR methods. Right: Different HR images generated by each method from an LR input crop of \emph{Forrest Gump}.}
\label{fig:conceptual_im}
\end{figure*}

\IEEEPARstart{S}{ingle-image} super-resolution (SISR) has long been a fundamental problem in low-level vision, aiming to recover a high-resolution (HR) image from an observed low-resolution (LR) input. Years of efforts from the research community have brought about remarkable progress in this field, especially with the booming deep learning techniques \cite{dong2014learning, dong2016accelerating, shi2016real, ledig2017photo, zhang2018image}. However, most existing methods assume a pre-defined degradation process (e.g., bicubic downsampling) from an HR image to an LR one, which can hardly hold true for real-world images with complex degradation types. Towards filling this gap, growing attention has been paid in recent years to approaches for unknown degradations, namely \emph{blind SR}. Despite many exciting improvements, these proposed methods tend to fail in many real-world scenarios, as their performance is usually limited to certain kinds of inputs and will drop dramatically in other cases. 
The main reason is that they still make some assumptions on the degradation types related to the input LR. Readers can see Fig.\ref{fig:conceptual_im}(a) for an illustration, which shows four different LR inputs with assumed degradation types of some state-of-the-art methods but targeting at the same HR. Therefore, when given an arbitrary input deviating from their assumed data distributions, these methods inevitably produce much less pleasing results. 
Fig.\ref{fig:conceptual_im}(b) demonstrates different SR results for a real-world image cropped from the famous film \emph{Forrest Gump}, which are generated by four state-of-the-art methods. We may find none of these methods have lived up to our expectation for a good viewing experience, since this real-world image does not strictly follow their assumptions on inputs.
In fact, it is not rare that we feel confused about which method to choose for a certain image at hand, or whether we can really get a high-quality result using existing methods.

In this paper, we try to relieve this confusion through a systematic survey on recent progress in blind SR with our own insight. 
What's more, it is highly necessary that we look back and reflect on the proposed methods to have a clear understanding of the current research state and remaining gaps. 
As stated above, we often have difficulty in selecting a proper method when facing so many ones: KernelGAN \cite{bell2019blind} for a single image looks cool, but how about IKC \cite{gu2019blind} with iterative scheme or CinCGAN \cite{yuan2018unsupervised} with unpaired training data? Also, even if every single blind SR method is claimed to work well for real images, we may still struggle to obtain a satisfactory output for our own image, just like the case in Fig.\ref{fig:conceptual_im}.
At this stage of development, it is time to ask: To what extent have we solved the problem? What is holding us back and where should we go for future endeavour? 

Hence, this paper aims to serve much more than a list of recent progress. Specifically, we propose a taxonomy to effectively categorize existing approaches, which clearly distinguishes among different methods and naturally reveals some research gaps.
Based on this taxonomy, our goal is to let each method have its own position within a broad picture composed of existing work. This picture can provide a guideline on reasonable and fair comparison between different kinds of methods in future work. 
In addition, we will make a summary on the application scopes along with limitations of each kind of approaches, helping readers to efficiently select appropriate methods for various scenarios.
Note that this paper focuses on SISR for general natural images, not including domain-specific topics like face SR or depth map SR. 

Our contributions are mainly three-fold: 1) We present a systematic survey on recent progress in blind image super-resolution, including the improvements and limitations of different approaches.
2) We propose a taxonomy to effectively categorize existing methods and reveal some research gaps. 
3) We provide our deep insight into current research state and promising future directions.

In the following sections, we first introduce the mathematical formulations of some commonly used SR models in Sec.\ref{sec:formulation}, and discuss about the challenges from real-world images that we face when addressing blind SR in Sec.\ref{sec:realworld_image_type}. Then we put forward our proposed taxonomy in Sec.\ref{sec:taxonomy}. A quick review on non-blind SISR is provided in Sec.\ref{sec:overview_SISR}, since the research state in non-blind setting has set up the foundation for blind SR. Then we elaborate on methods of each category in Sec.\ref{sec:one} and Sec.\ref{sec:two}, followed by a summary on commonly used datasets and previous competitions in the field of blind SR in Sec.\ref{sec:dataset_competition}. Quantitative and qualitative comparison among some representative methods is included in Sec.\ref{sec:result}. Finally, we draw a conclusion on our insight through this survey and our perspective on future directions in Sec.\ref{sec:conclusion}.

\section{Problem Formulation}\label{sec:formulation}

\begin{figure}[ht]
\renewcommand{\captionfont}{\small \rmfamily}
\begin{center}
\includegraphics[width=0.9\linewidth]{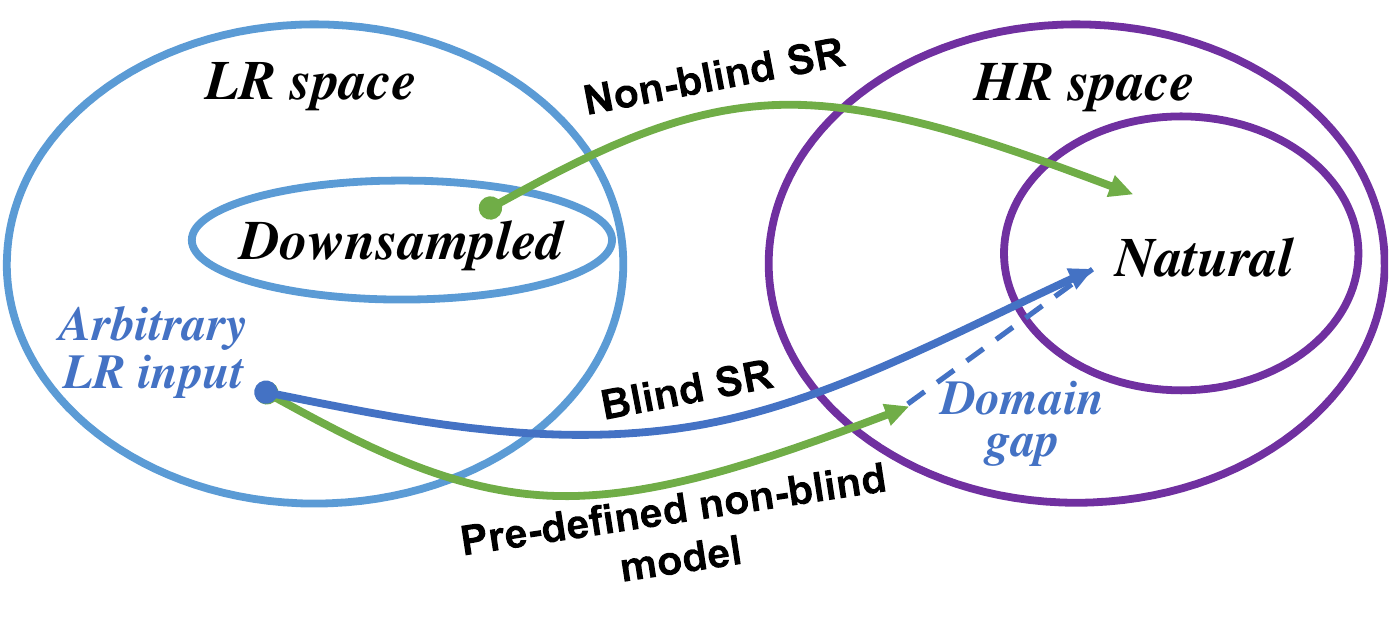}
\end{center}
\vspace{-12pt}
\caption{Domain interpretation of differences between non-blind and blind SR. There exists a large domain gap between the SR result and desired high-quality HR, which is caused by applying a pre-trained non-blind model to LR input with degradation deviating from the assumed one (e.g., downsampling).}
\label{fig:domain_pic}
\end{figure}

\begin{figure*}[ht]
\renewcommand{\captionfont}{\small \rmfamily}
\begin{center}
   \includegraphics[width=1.0\linewidth]{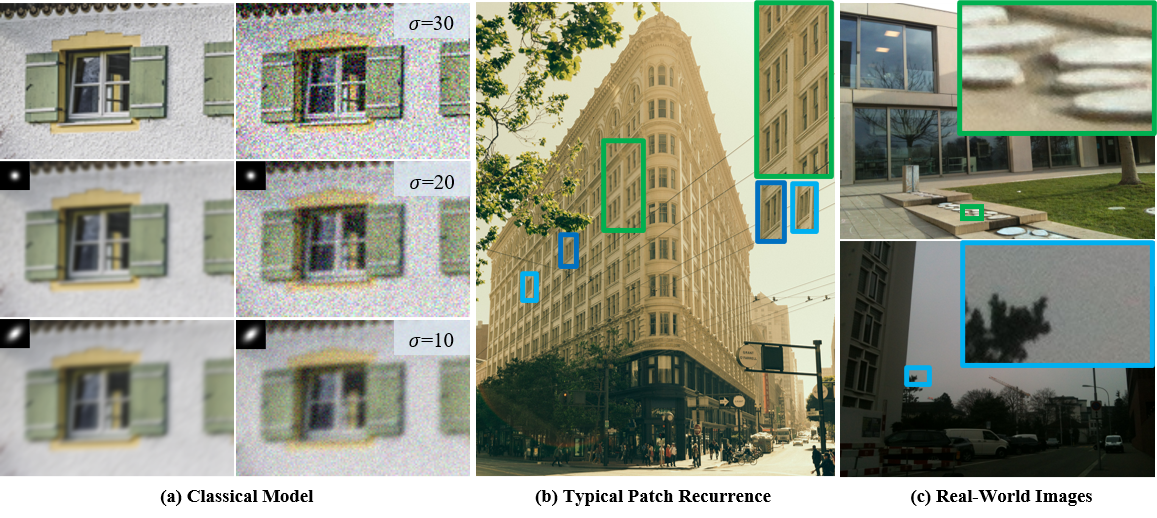}
\end{center}
\vspace{-12pt}
   \caption{Examples of degraded LR image. (a) Clean LR (top left, generated with bicubic downsampling) and its blurry or noisy versions with different $\boldsymbol{k}$ and $\boldsymbol{n}$. The 2$^{nd}$ row is with isotropic Gaussian kernels while the 3$^{rd}$ row with anisotropic Gaussian, and $\sigma$ is variance of additive Gaussian noise. The example image is from DIV8K \cite{DBLP:conf/iccvw/GuLDFLT19}; (b) Image with typical patch recurrence, both within and across different scales of the same image. The image is from Urban100 \cite{DBLP:conf/cvpr/HuangSA15}; (c) Real-world smartphone images with complex unknown degradations. Images are from DPED dataset \cite{DBLP:conf/iccv/IgnatovKTVG17}.}
\label{fig:real_lr_im}
\end{figure*}

In this section, we introduce some mathematical formulations of the SISR problem. Specifically, SISR refers to the task of reconstructing an HR image from a given LR input, especially the high-frequency contents in HR. The underlying degradation process from HR to LR can be generally expressed with the following equation:
\begin{equation}
    \label{equ:abstract_model}
    \centering
    \boldsymbol{y}=f(\boldsymbol{x}; s),
\end{equation} 
where $\boldsymbol{x}$, $\boldsymbol{y}$ denote HR image and LR image respectively, \emph{f} is the degradation function with a scale factor \emph{s}. 
Therefore, SR problem is equivalent to modelling and solving the inverse function \emph{f$^{-1}$}.
In the background of \emph{non-blind SR}, \emph{f} is usually assumed to be bicubic downsampling:
\begin{equation}
    \label{equ:bicubic_model}
    \centering
    \boldsymbol{y}=\boldsymbol{x}\downarrow_s^{bic},
\end{equation} 
or the combination of downsampling and a fixed Gaussian blur with kernel $\boldsymbol{k}_g$:
\begin{equation}
    \label{equ:gauss_bic_model}
    \centering
    \boldsymbol{y}=(\boldsymbol{x}\otimes \boldsymbol{k}_{g})\downarrow_s,
\end{equation} 
where $\otimes$ denotes convolutional operation. Under either assumption, the corresponding SR model is only able to handle LR inputs with this specific kind of degradation. For other LR images with different degradation types, the mismatch between SR model and intrinsic degradation of inputs may cause severe artifacts in SR results \cite{gu2019blind,shocher2018zero}. 
Fig.\ref{fig:domain_pic} gives an illustration on this mismatch from the perspective of image domain adaptation: if an SR model corresponding to a pre-defined degradation is applied to an arbitrary LR input, there will be a large domain gap between the SR output and desired image samples from the target \emph{Natural HR} domain, thus leading to a poor-quality result.

Hence, the topic of blind SR for unknown degradation is proposed in an attempt to bridge this gap. Up till now, there have been two different ways of modelling the degradation process for blind SR: explicit modelling based on an extension of Equation (\ref{equ:gauss_bic_model}), and implicit modelling through inherent distribution within external dataset.

To be specific, explicit modelling usually employs a so-called classical degradation model, which is a more general form of Equation (\ref{equ:gauss_bic_model}): 
\begin{equation}
    \label{equ:kernel_model}
    \centering
    \boldsymbol{y}=(\boldsymbol{x}\otimes \boldsymbol{k})\downarrow_s+\boldsymbol{n},
\end{equation} 
where the SR blur kernel $\boldsymbol{k}$ and additive noise $\boldsymbol{n}$ are two main factors involved in the degradation process, and parameters related to these two factors will be unknown for an arbitrary LR input. 
Fig.\ref{fig:real_lr_im}(a) shows several image examples with different $\boldsymbol{k}$ and $\boldsymbol{n}$, which are much more degraded than their bicubic-downsampled counterpart.
Some approaches utilize external dataset to learn an SR model well adapted to a large set of various $\boldsymbol{k}$ or $\boldsymbol{n}$, such as IKC \cite{gu2019blind} and SRMD \cite{zhang2018learning}. 
Besides blurring and noise, more complex and realistic degradation types can also be involved into the formulation, like JPEG compression with a quality factor \emph{q} \cite{liu2020learning}:
\begin{equation}
    \label{equ:jpeg_model}
    \centering
    \boldsymbol{y}=((\boldsymbol{x}\otimes \boldsymbol{k})\downarrow_s+\boldsymbol{n})_{JPEG_{q}},
\end{equation} 
Another group of methods leverage the internal statistics within a single image derived from the classical degradation model, thus requiring no external dataset for training, like ZSSR \cite{shocher2018zero} and DGDML-SR\cite{chengzero}. In fact, internal statistics just reflects the patch recurrence property of an image, and readers can refer to Fig.\ref{fig:real_lr_im}(b) for an illustration.

Nevertheless, real-world degradations are usually too complex to be modelled with an explicit combination of multiple degradation types, as shown in Fig.\ref{fig:real_lr_im}(c). Therefore, implicit modelling attempts to circumvent the explicit modelling function. Instead, it defines the degradation process \emph{f} implicitly through data distribution, and all the existing approaches with implicit modelling require an external dataset for training. Typically, these methods utilize data distribution learning with Generative Adversarial Network (GAN) \cite{goodfellow2014generative} to grasp the implicit degradation model possessed within training dataset, like CinCGAN \cite{yuan2018unsupervised}.

Although so many models have been put forward in blind SR, there is still a long way to go since we have only tackled a small set of real-world images. Existing methods often claim to focus on real-world settings, but they actually assume a certain scene, like images taken by some digital cameras \cite{lugmayr2020ntire, DBLP:conf/eccv/WeiLTLZPLXFZLHD20}.
In fact, real-world images are greatly different in their underlying degradation types, and an SR model designed for a specific type can easily fail for another.
In the next section, we will give a brief discussion on different kinds of real-world images, which have posed severe challenges to the field of blind SR.

%------------------------------------------------------------------
\section{Challenges from Real-World Images}\label{sec:realworld_image_type}

\begin{figure}[htbp]
\renewcommand{\captionfont}{\small \rmfamily}
\begin{center}
\includegraphics[width=1.0\linewidth]{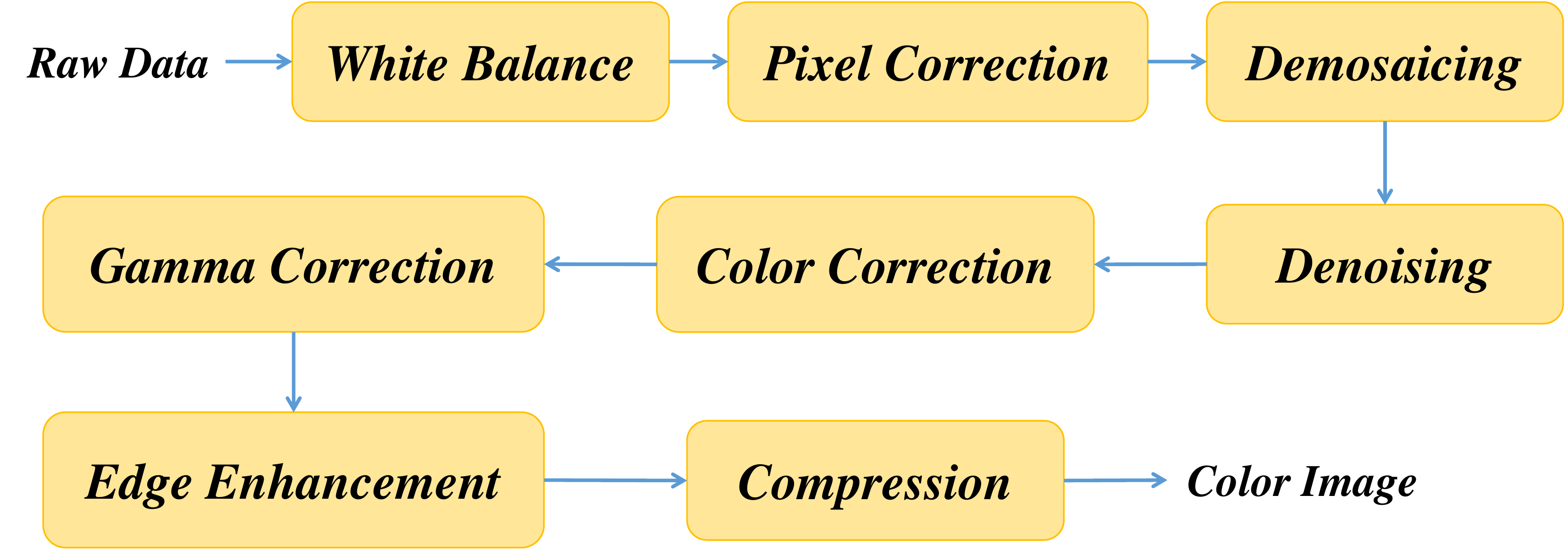}
\end{center}
\vspace{-8pt}
\caption{Image signal processing pipeline.}
\label{fig:real_world_im_class_2}
\end{figure}

\begin{figure}[htbp]
\renewcommand{\captionfont}{\small \rmfamily}
\begin{center}
\includegraphics[width=1.0\linewidth]{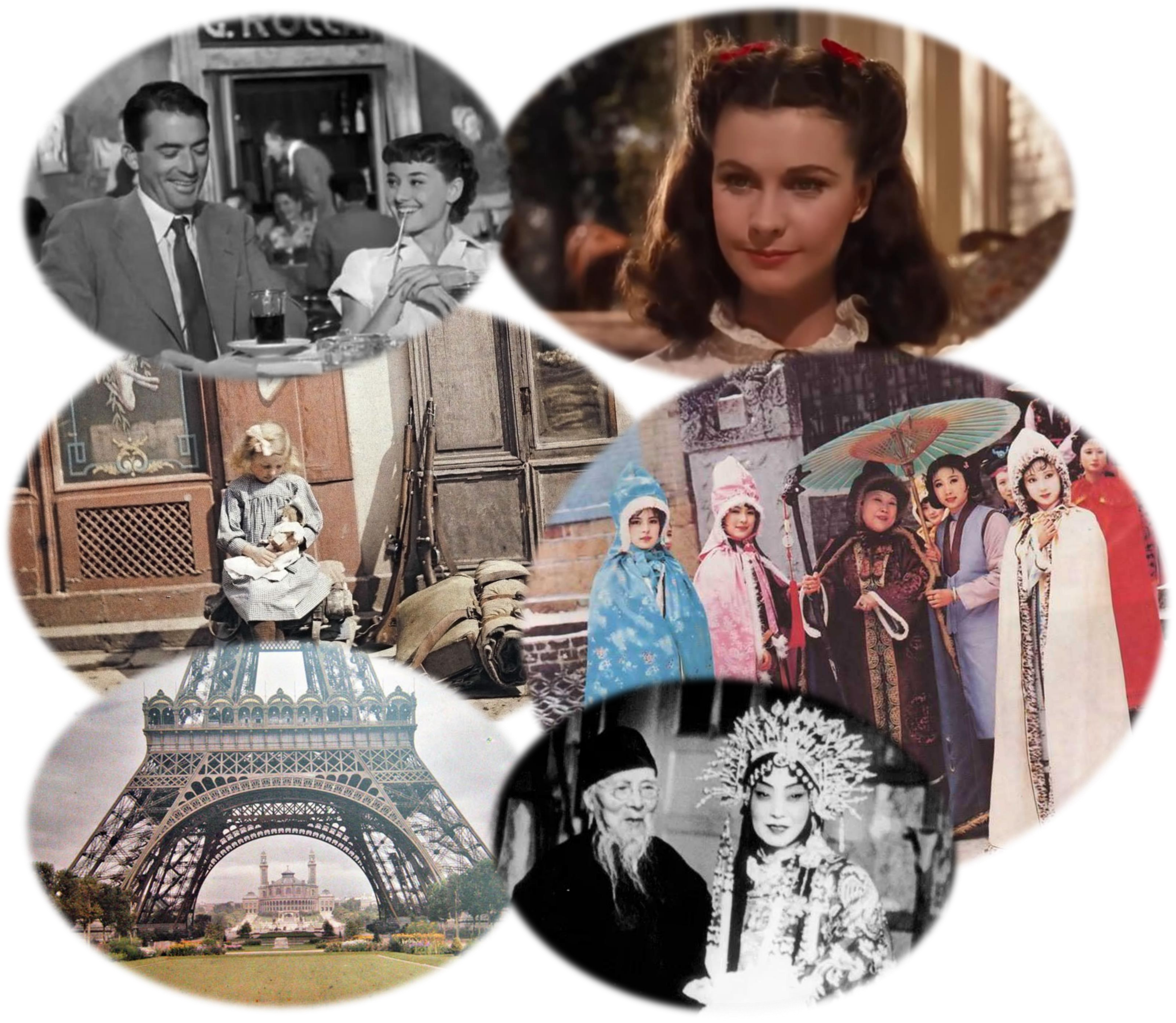}
\end{center}
\vspace{-12pt}
\caption{Old photos with degradation rising from storage.}
\label{fig:real_world_im_class_1}
\end{figure}

With the development of modern imaging devices, we are now embracing the world with a surge of visual data. Such a variety of image sources also pose more challenges, especially in terms of degradations types.
Generally speaking, there are three main factors causing different degradations:

\begin{itemize}
\item[(1)] {Different imaging devices. This era of technology has given birth to a dazzling array of digital cameras, not to mention smartphones with advanced camera systems. However, these devices differ greatly in the characteristics of the photos taken \cite{DBLP:conf/iccv/IgnatovKTVG17}. For example, a DSLR (digital single-lens reflex) camera is able to capture high-quality images with a sense of stereoscopy by adjusting its focal length, while a smartphone camera is nowhere near DSLR-quality, tending to produce a "flattened" and noisy scene due to its physical limitations in sensor size and lenses. Another type of low-quality imaging is surveillance video, which often suffers greatly from loss of focus. Readers can see Fig.\ref{fig:real_world_im_class} for some image examples. Images captured with different devices can thus have distinct degradations from one another.}

\begin{figure}[htbp]
\renewcommand{\captionfont}{\small \rmfamily}
\begin{center}
\includegraphics[width=0.98\linewidth]{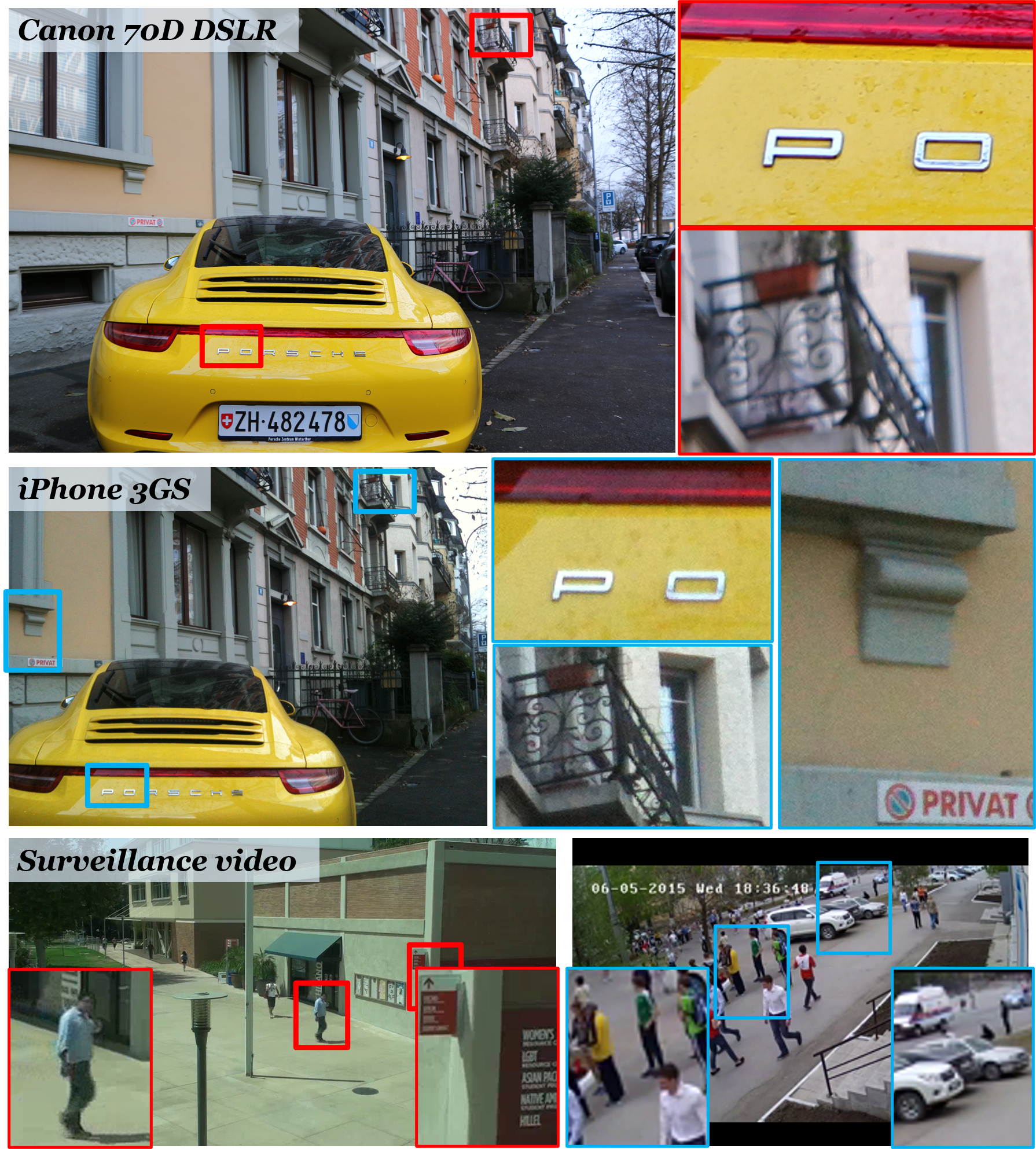}
\end{center}
\vspace{-10pt}
\caption{Real-world images (or video frames) captured with different devices. Images are from DPED \cite{DBLP:conf/iccv/IgnatovKTVG17}, VIRAT \cite{DBLP:conf/cvpr/OhHPCCLMALDSWJRSVPRYTSFRD11}, UCF-Crime \cite{DBLP:conf/cvpr/SultaniCS18} datasets.}
\label{fig:real_world_im_class}
\end{figure}

\item[(2)] {Image processing algorithms. This problem is mostly related to digital and smartphone cameras, since it is an image signal processor on the chip that actually processes digital signals into images. The processing pipeline usually involves multiple steps, such as pixel correction, white balance correction, denoising and sharpening. During this process, complex unknown degradations can be introduced \cite{qian2019trinity}, which are unpredictable and varying among different devices. A typical pipeline is shown in Fig.\ref{fig:real_world_im_class_2}.}

\item[(3)] {Degradations rising from storage. To reduce the resource consumption for transmitting and storing data, images and videos are always compressed. Accompanying compressed images are compression artifacts, which will lead to degradations like blurring and blocky effects. In addition, time itself can gradually deteriorate images, especially for old photos and movies recorded on films. Such degradations are mainly caused by poor imaging equipments or erosion in the air, including film grain, sepia effect and color fading \cite{DBLP:journals/corr/abs-2009-07047}. Some example images are presented in Fig.\ref{fig:real_world_im_class_1}. This kind of degradations can hardly be expressed with explicit functions or covered by a few external datasets, thus demanding more efforts in designing restoration algorithms.}
\end{itemize}

The real-world images discussed above all bear their own degradations and challenges. Nonetheless, previous work usually focus on a single type of real images, like those taken by smartphones, which greatly limits their performance in diverse scenes. 
We expect to see more explorations on different types of real-world images in the future. Specifically, effective solutions for each distinct type, if not a general solution to all, should be the ultimate goal of our research community.

%------------------------------------------------------------------
\section{Taxonomy}\label{sec:taxonomy}

In this section, we will elaborate on our proposed taxonomy to serve as the guideline for our review and analysis. According to Sec.\ref{sec:formulation}, there have been two ways of modelling the degradation process involved in blind SR: explicit modelling based on the classical degradation model or its variants, and implicit modelling using data distribution among external dataset. 
The basic idea of explicit modelling is to learn an SR model with external training data covering a large set of degradations, which are usually parameterized with $\boldsymbol{k}$ and $\boldsymbol{n}$ in Equation (\ref{equ:kernel_model}). Representative approaches include SRMD \cite{zhang2018learning}, IKC \cite{gu2019blind} and KMSR \cite{zhou2019kernel}. 
Another group of methods propose to exploit internal statistics of patch recurrence, like KernelGAN \cite{bell2019blind} and ZSSR \cite{shocher2018zero}, which can directly work on a single input image. This kind of modelling is primarily based on the classical degradation model.
On the other hand, methods with implicit modelling do not depend on any explicit parameterization, and they usually learn the underlying SR model implicitly through data distribution within external dataset. Among such methods are CinCGAN \cite{yuan2018unsupervised} and FSSR \cite{fritsche2019frequency}.

\begin{figure}[htbp]
\renewcommand{\captionfont}{\small \rmfamily}
\begin{center}
\includegraphics[width=1.0\linewidth]{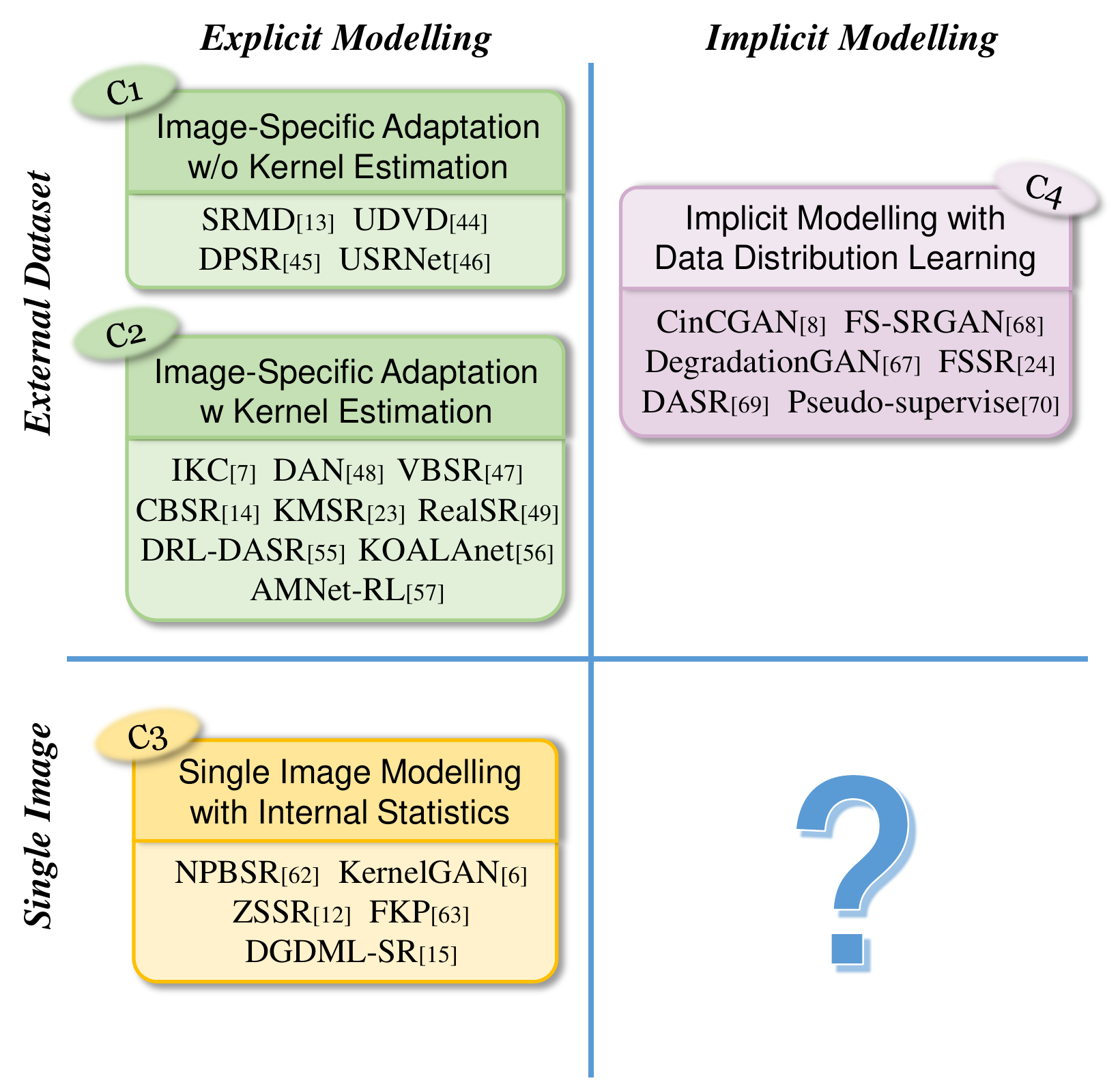}
\end{center}
\vspace{-10pt}
\caption{Our proposed taxonomy and the corresponding representative methods. Our taxonomy distinguishes the ways of degradation modelling and data used for solving SR models, which also naturally reveals the remaining research gap.}
\label{fig:classes}
\end{figure}

Therefore, we propose a taxonomy to effectively classify existing approaches according to their ways of degradation modelling and the used data for solving the SR model: explicit modelling or implicit modelling, external dataset or a single input image, as shown in Fig.\ref{fig:classes}. Our reasons for adopting this categorization are three-fold: first, distinguishing between explicit and implicit modelling helps us to understand the assumption of a certain method, i.e., what kind of degradations this method aims to deal with; second, whether using external dataset or a single input image indicates different strategies of image-specific adaptation with explicit modelling; finally, after categorizing existing approaches into these classes, one remaining research gap naturally reveals itself - \emph{implicit modelling with a single image}. We argue that this direction is promising in terms of addressing general real-world images with diverse contents, and we will also try to propose feasible suggestions for new solutions in this direction. 

In the next sections, we first give a quick overview on non-blind SISR, which sets the basis for blind SR methods. Then methods with explicit modelling are introduced in Sec.\ref{sec:one}, and those using implicit modelling are discussed in Sec.\ref{sec:two}. 
For each type of methods, we will unfold the review along its \emph{\textbf{course of development}}, and make an analysis on their limitations to inspire future work.

%-------------------------------------------------------------------------
\section{Overview of Non-Blind Single-Image Super-Resolution}\label{sec:overview_SISR}

\begin{figure}[htbp]
\renewcommand{\captionfont}{\small \rmfamily}
\begin{center}
\includegraphics[width=0.95\linewidth]{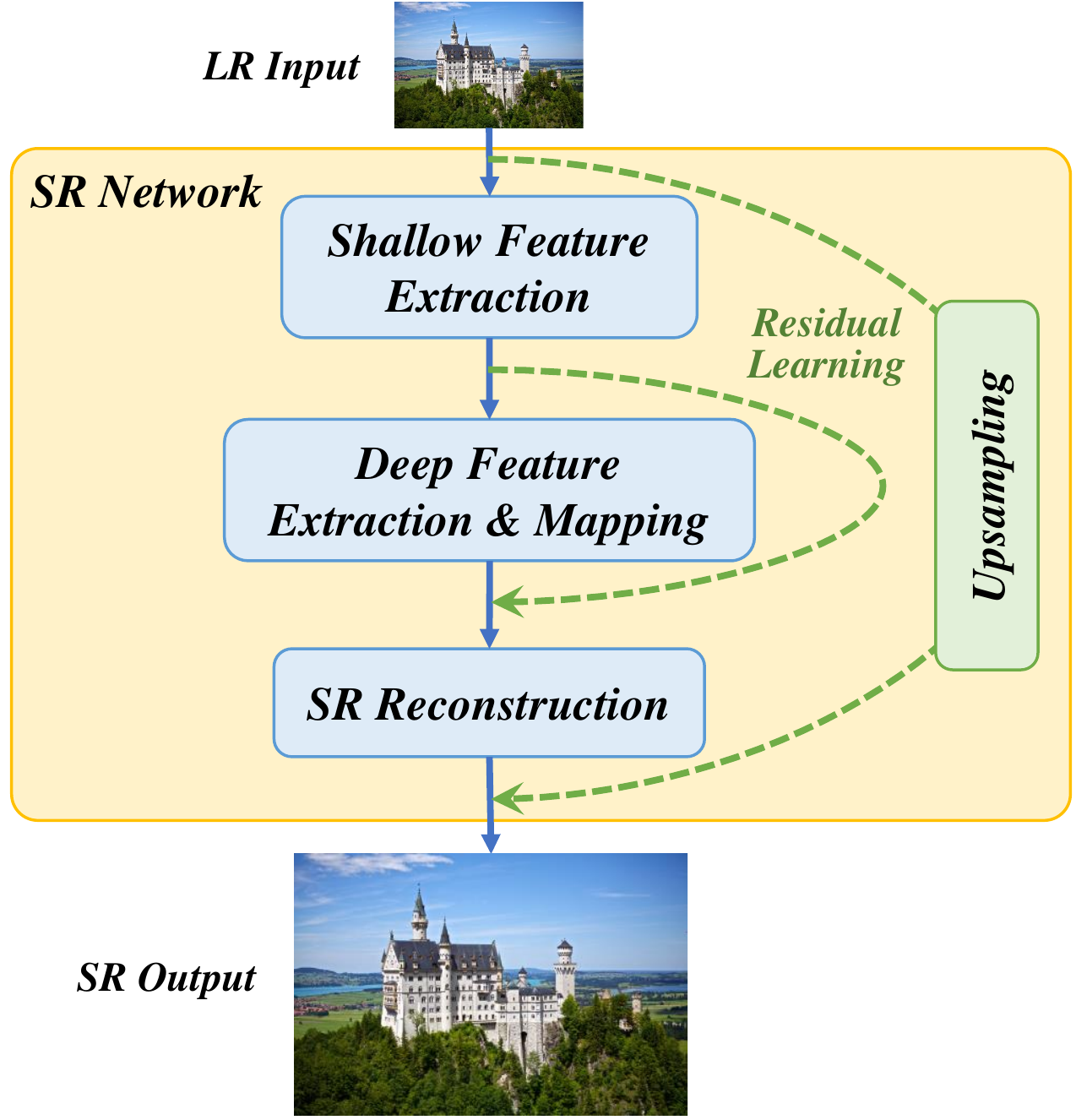}
\end{center}
\vspace{-8pt}
\caption{Common CNN framework of non-blind SISR.}
\label{fig:nonblind_frame}
\end{figure}

As explained in Sec.\ref{sec:formulation}, non-blind SR assumes a fixed known degradation process for solving HR outputs. Before the development of deep learning techniques, many traditional techniques are example-based. \cite{DBLP:conf/cvpr/ChangYX04, DBLP:conf/cvpr/YangWHM08, DBLP:conf/iccv/TimofteDG13, DBLP:journals/tip/YangWHM10} learn the mapping function from LR to HR with external HR-LR exemplar pairs, where the mapping learning is usually based on a compact dictionary or manifold space. Some others \cite{DBLP:journals/tog/FreedmanF11, DBLP:conf/iccv/GlasnerBI09} utilize the property of internal self-similarity within a single image without employing external dataset. In 2014, the pioneering work of SRCNN \cite{DBLP:conf/eccv/DongLHT14} opened a new era of deploying convolutional neural network (CNN) to tackle this task, and it also set up the basic framework for later works, as shown in Fig.\ref{fig:nonblind_frame}. 

\begin{figure*}[htbp]
\renewcommand{\captionfont}{\small \rmfamily}
\begin{center}
\includegraphics[width=1.0\linewidth]{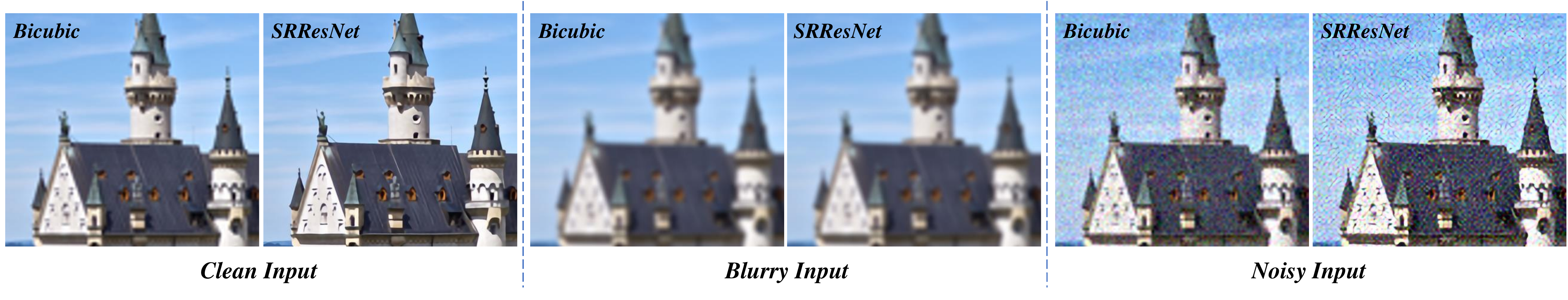}
\end{center}
\vspace{-10pt}
\caption{Failure cases of non-blind SR network, e.g. SRResNet \cite{DBLP:conf/cvpr/LedigTHCCAATTWS17}. Compared with results generated by bicubic interpolation, SRResNet recovers little sharp texture for the blurry input, but also unfavourably keeps the noises for the noisy input.}
\label{fig:nonblind_fail}
\end{figure*}

The commonly adopted CNN framework for SISR task includes three main modules: shallow feature extraction to convert an input LR image into feature maps, deep feature extraction or mapping based on extracted shallow features, and finally SR output reconstruction. Residual learning has also been widely adopted to ease the training process, either in image-level \cite{vdsr} or feature-level \cite{esrgan}. 
Recent years have witnessed many improvements on deep feature extraction and SR reconstruction modules, such as introducing residual blocks \cite{DBLP:conf/cvpr/LimSKNL17, DBLP:conf/iccv/0001LLG17, esrgan}, recursive or recurrent structure \cite{DBLP:conf/cvpr/KimLL16, DBLP:conf/cvpr/TaiY017}, attention mechanism \cite{DBLP:conf/eccv/ZhangLLWZF18, DBLP:conf/cvpr/DaiCZXZ19}, sub-pixel convolution \cite{DBLP:conf/cvpr/ShiCHTABRW16}, etc. 
In addition, multiple loss functions are also proposed for better perceptual quality of SR results \cite{DBLP:conf/cvpr/LedigTHCCAATTWS17, DBLP:conf/iccv/SajjadiSH17, DBLP:conf/iccv/ZhangLDQ19}. 
These techniques bring about remarkable progress in terms of both reconstruction accuracy and efficiency, and non-blind SISR with bicubic-downsampling assumption actually reaches maturity.

However, these non-blind models usually struggle to generalize to input images with more complex degradations deviating from their assumed one. 
Some failure cases of a non-blind SR network are shown in Fig.\ref{fig:nonblind_fail}, where the network performs well on bicubicly downsampled clean input in accordance with the assumed degradation model, but cannot handle blurry or noisy input images.
Hence, it is demanding to propose methods for blind SR setting, which are the main focus of this survey and will be explored in detail in the following two sections.

%-------------------------------------------------------------------------
\section{Explicit Degradation Modelling}\label{sec:one}
This section covers recently proposed blind SR methods with explicit modelling of degradation process, usually based on the classical degradation model shown by Equation (\ref{equ:kernel_model}). What's more, these approaches can be further classified into two sub-classes according to whether they employ external dataset or rely on a single input image to solve the SR problem.

%%-------------------------------------------------------------------------

\begin{figure}[htbp]
\renewcommand{\captionfont}{\small \rmfamily}
\begin{center}
\includegraphics[width=0.98\linewidth]{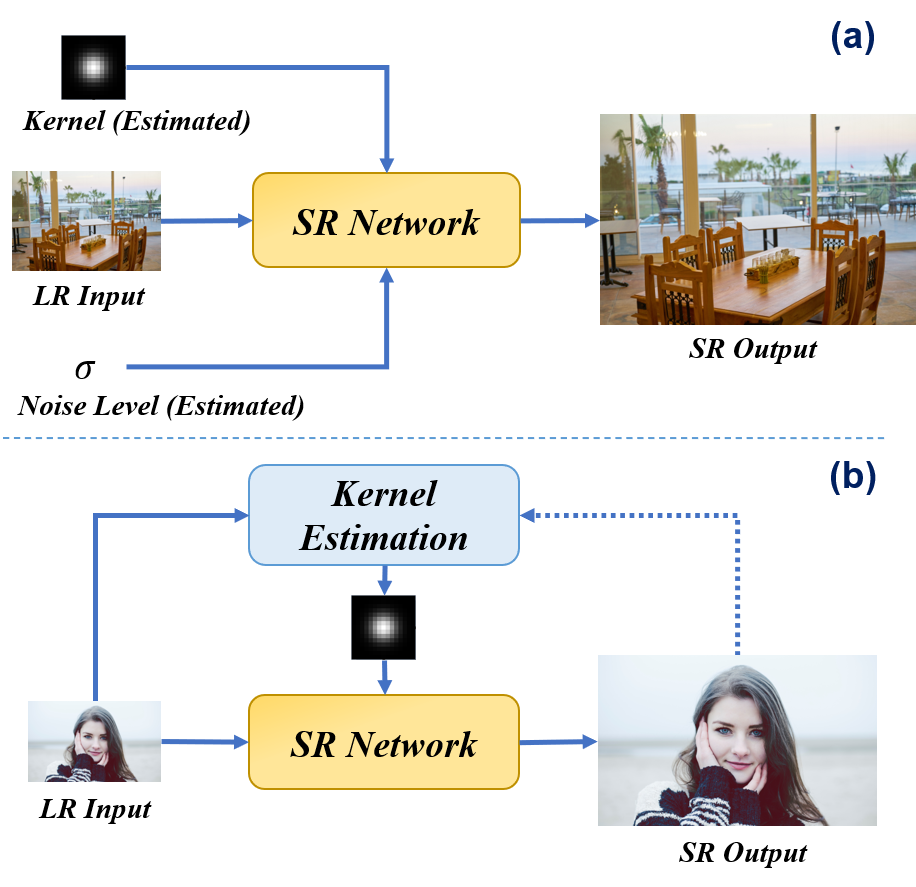}
\end{center}
\vspace{-12pt}
\caption{Two types of overall frameworks for methods with explicit modelling and external dataset. (a) Image-specific adaptation \emph{without} kernel estimation; (b) Image-specific adaptation \emph{with} kernel estimation. The connection with dotted line indicates that it is optional.}
\label{fig:1_1_struct}
\end{figure}

\subsection{Classical Degradation Model with External Dataset}\label{sec:one_one}

This kind of approaches utilize external dataset to train an SR model well adapted to variant SR blur kernels $\boldsymbol{k}$ and noises $\boldsymbol{n}$, especially the former. Typically, the SR model is parameterized with a convolutional neural network (CNN), and an estimation on $\boldsymbol{k}$ or $\boldsymbol{n}$ for a specific LR image is used as conditional input to the SR model for feature adaptation purpose. After the training process, the model will be able to produce satisfactory results for LR inputs with degradation types covered in the training dataset. According to whether a certain approach includes degradation estimation in its proposed framework, we further divide these approaches into two types: image-specific adaptation \emph{without} kernel estimation, and image-specific adaptation \emph{with} kernel estimation. To be more specific, the first type receives estimated degradation information as additional inputs and is focused on how to utilize the estimation input for image-specific adaptation, while the second one pays special attention to kernel estimation along with the SR process. An illustration of their overall frameworks is presented in Fig.\ref{fig:1_1_struct}.

%%-------------------------------------------------------------------------
\subsubsection{Image-Specific Adaptation without Kernel Estimation}\label{sec:one_one_one}

\begin{figure*}[htbp]
\renewcommand{\captionfont}{\small \rmfamily}
\begin{center}
\includegraphics[width=0.98\linewidth]{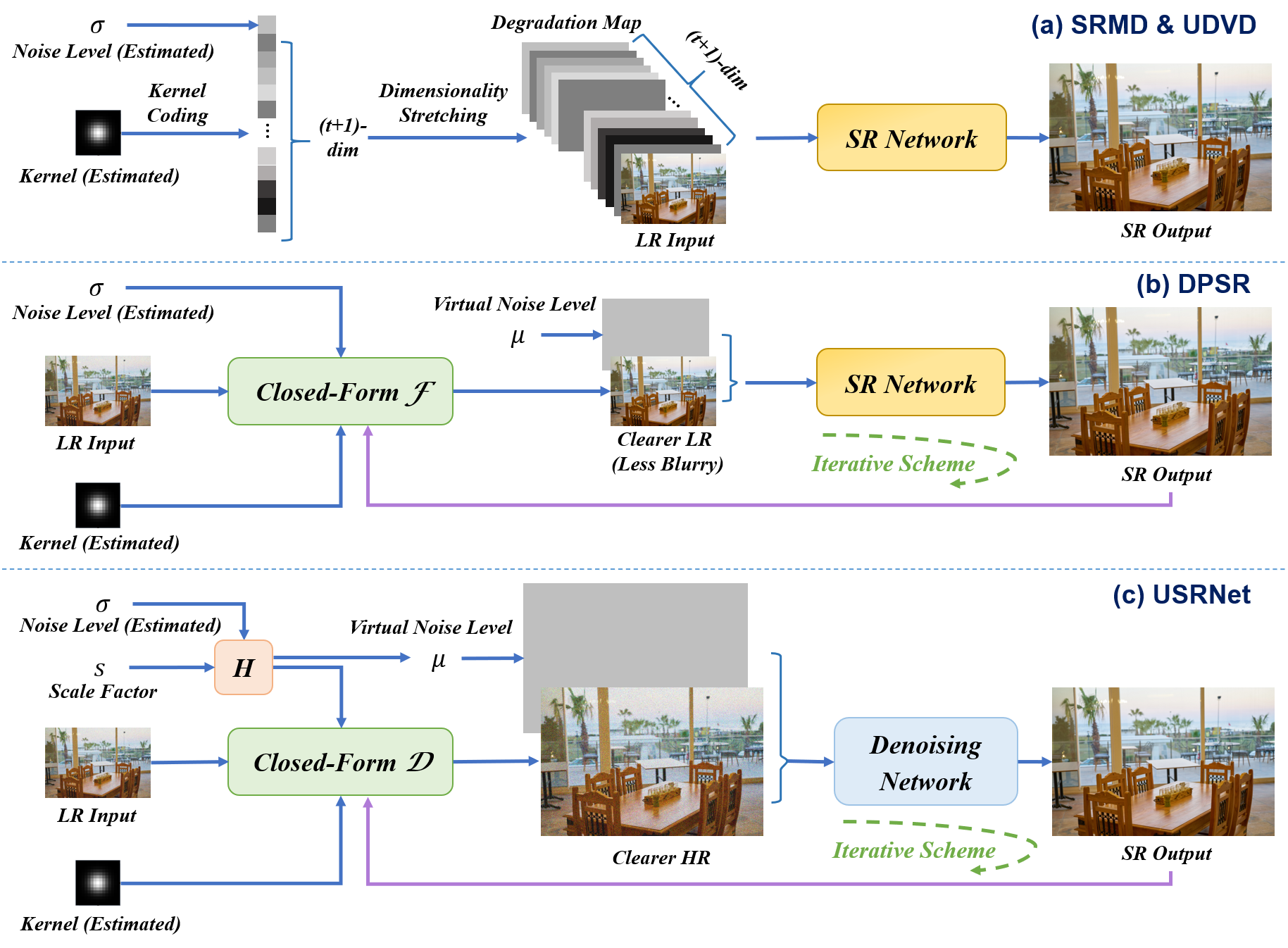}
\end{center}
\vspace{-10pt}
\caption{Pipeline of methods in \emph{\textbf{Image-Specific Adaptation without Kernel Estimation}}. (a) SRMD \cite{zhang2018learning} and UDVD \cite{xu2020unified}; (b) DPSR \cite{Zhang_2019_CVPR}; (c) USRNet \cite{DBLP:conf/cvpr/ZhangGT20}.}
\label{fig:3_1_1_struct}
\end{figure*}

\begin{figure*}[htbp]
\renewcommand{\captionfont}{\small \rmfamily}
\begin{center}
\includegraphics[width=1.0\linewidth]{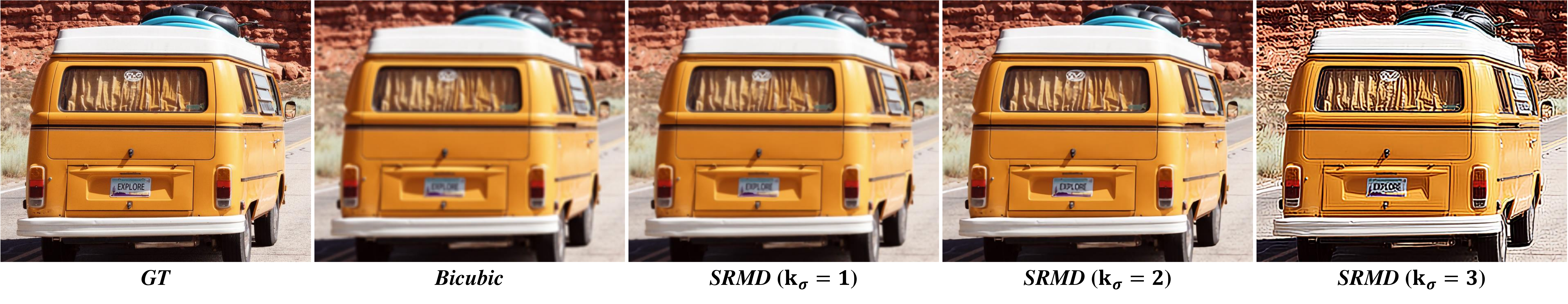}
\end{center}
\vspace{-12pt}
\caption{Illustration on the limitations of \emph{\textbf{Image-Specific Adaptation without Kernel Estimation}}. Input image is blurred and downsampled with Gaussian kernel $\boldsymbol{k}_{\sigma}$=2. An incorrect kernel estimation input either leads to blurry output or unnatural ringing artifacts with over-enhanced textures.}
\label{fig:limit_3_1_1}
\end{figure*}

Super-resolution for multiple degradations (SRMD) \cite{zhang2018learning} proposes to directly concatenate an LR input image with its degradation map as a unified input to the SR model, thus allowing feature adaptation according to the specific degradation and covering multiple degradation types in a single model. In order to generate degradation map with the same dimension as the LR image, it introduces a strategy called dimensionality stretching. Specifically, an SR blur kernel with size \emph{r$\times$r} is flattened to an \emph{r}$^{2}$-length vector and reduced to \emph{t}-dim with principal component analysis (PCA) to get the kernel coding. After concatenating with the estimated noise level $\boldsymbol{\sigma}$ related to $\boldsymbol{n}$, the \emph{(t+1)}-dim vector is stretched both vertically and horizontally to get the final \emph{H$\times$W$\times$(t+1)}-dim degradation map, where \emph{H} and \emph{W} are height and width of the LR image. This strategy can be easily extended to non-uniform maps for spatially variant degradations. The SR reconstruction network of SRMD is similar to those commonly adopted in non-blind SR. 
The whole pipeline is presented in Fig.\ref{fig:3_1_1_struct}(a).

Following SRMD, UDVD \cite{xu2020unified} also uses the degradation map as an additional input for SR reconstruction, yet it makes one step forward by employing per-pixel dynamic convolution to more effectively deal with variational degradations across images. Specifically, a refinement network composed of several dynamic blocks with dynamic convolution is cascaded to feature extraction module, and each of these blocks refines SR output in an iterative way based on the result of its previous block. In addition, an improvement on the kernel coding operation is proposed by variant blind SR \cite{cornillere2019blind} to replace the PCA technique with a shallow neural network, which can potentially learn a kernel mapping more fitted to the specific SR model.

\begin{figure*}[htbp]
\renewcommand{\captionfont}{\small \rmfamily}
\begin{center}
\includegraphics[width=0.98\linewidth]{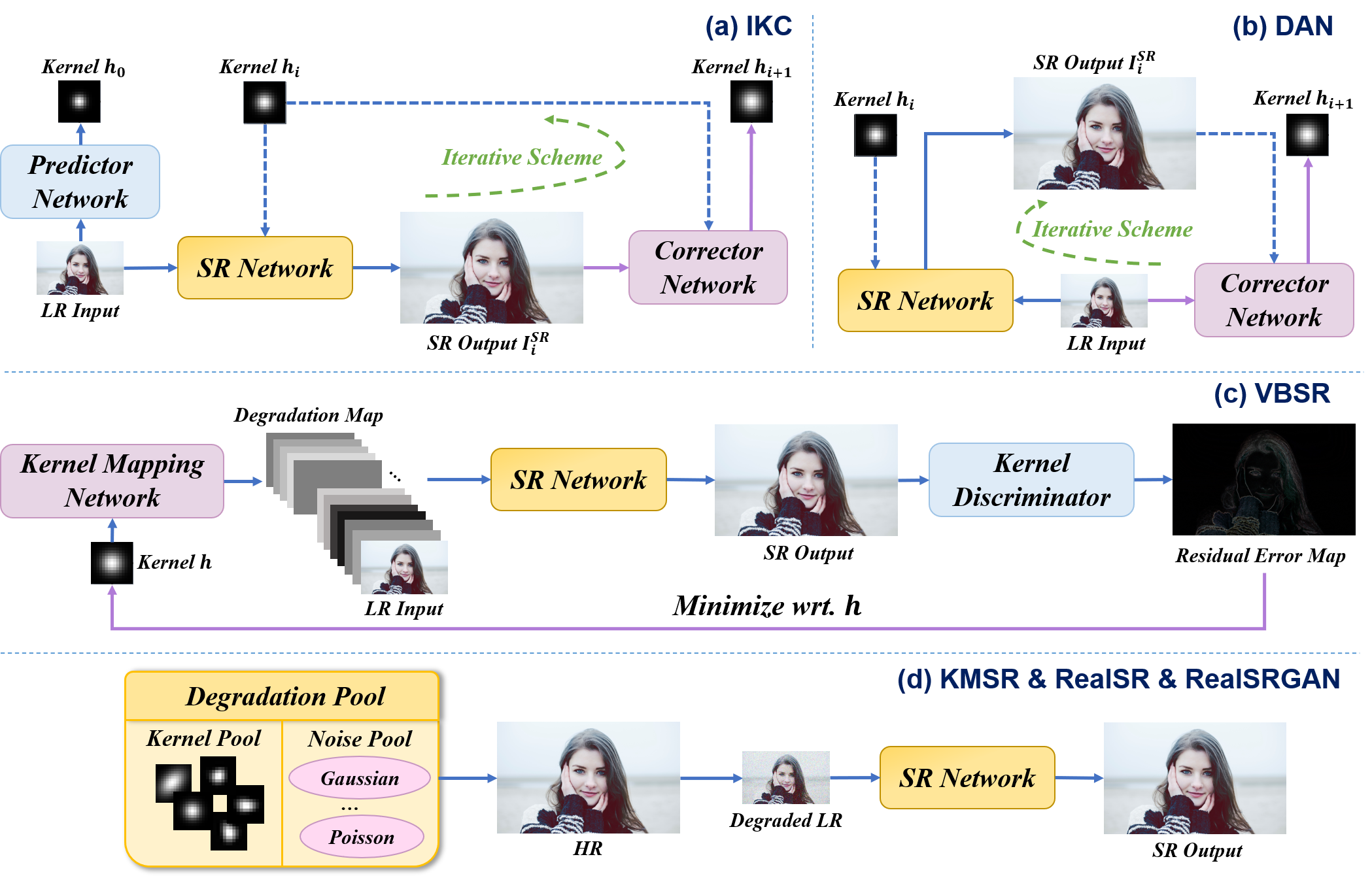}
\end{center}
\vspace{-5pt}
\caption{Detailed frameworks of methods from \emph{\textbf{Image-Specific Adaptation with Kernel Estimation}}. (a) IKC \cite{gu2019blind}; (b) DAN \cite{huang2020unfolding}; (c) VBSR \cite{cornillere2019blind}; (d) KMSR \cite{zhou2019kernel}, RealSR \cite{Ji_2020_CVPR_Workshops} and RealSRGAN \cite{ren2020real}. A connection with dotted line denotes a conditional input.}
\label{fig:3_1_2_struct}
\end{figure*}

\begin{figure*}[htbp]
\renewcommand{\captionfont}{\small \rmfamily}
\begin{center}
\includegraphics[width=1.0\linewidth]{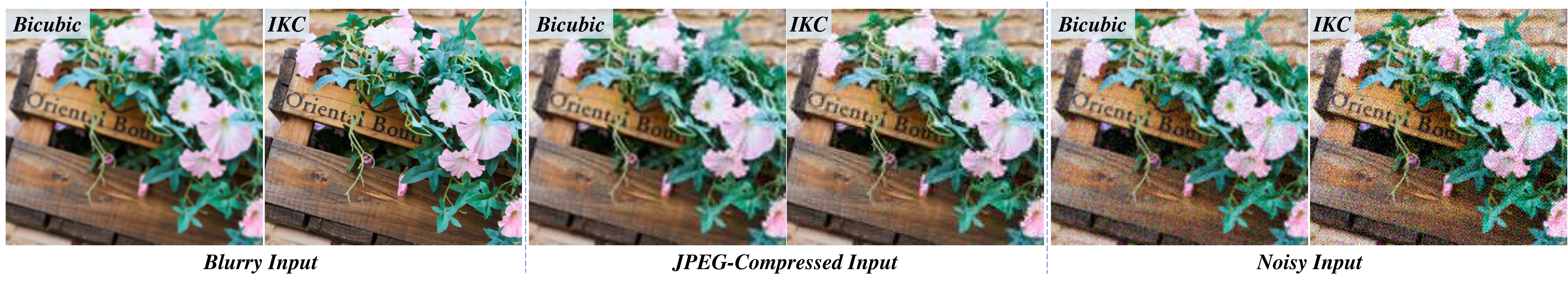}
\end{center}
\vspace{-12pt}
\caption{Illustration on the limitations of \emph{\textbf{Image-Specific Adaptation with Kernel Estimation}}. For input images with degradation types not covered in the SR model, like JPEG compression and Gaussian noise for IKC \cite{gu2019blind}, the obtained SR results deteriorate dramatically. Best viewed on screen.}
\label{fig:limit_3_1_2}
\end{figure*}

Although SRMD extends the generalization capacity of an SR model to variant SR kernels and noise levels, it still has very limited scope since it is usually non-trivial to effectively encode an arbitrary kernel and handle it with a single model, especially for those with irregular patterns like motion blur. 
Hence, another group of methods have been proposed based on an MAP framework, which requires no kernel coding for degradation map generation. Specifically, deep plug-and-play super-resolution (DPSR) \cite{Zhang_2019_CVPR} incorporates SR network into an MAP-based iterative optimization scheme. It primarily solves the HR image by minimizing the following objective function, which consists of a data term \emph{\textbf{D}} and a prior term \emph{\textbf{P}} regularized by a parameter $\lambda$: 
\begin{equation}
    \label{equ:MAP_objective}
    \centering
    E(\boldsymbol{x})=\frac{1}{2\sigma ^{2}}\left \| \boldsymbol{y}-\boldsymbol{x}\downarrow_{s}\otimes \boldsymbol{k} \right \|^{2}+\lambda \Phi (\boldsymbol{x})=D + \lambda P,
\end{equation}
whose corresponding degradation model is a modified version of Equation (\ref{equ:kernel_model}), decoupling the downsampling process from blurring operation:
\begin{equation}
    \label{equ:kernel_model_}
    \centering
    \boldsymbol{y}=(\boldsymbol{x}\downarrow_s\otimes \boldsymbol{k})+\boldsymbol{n}.
\end{equation} 
The objective function shown by Equation (\ref{equ:MAP_objective}) can be split into two sub-problems using the half-quadratic splitting (HQS) algorithm: one addresses deblurring task and is related to data term \emph{\textbf{D}} with parameter $\boldsymbol{k}$, while the other aims to super-resolve a bicubic downsampled image with some virtual noise level \emph{$\mu$} and is related to prior term \emph{\textbf{P}}. Fortunately, the first sub-problem can be solved in a closed form with Fast Fourier Transform without any kernel coding, thus allowing the model to cope with more complex kernels. Moreover, thanks to the decoupling of blurring and downsampling operations, the second sub-problem can be modelled by a non-blind SR network capable of dealing with additive noise, and this network can be directly adapted from SRMD framework with a single noise map as additional input.
Unfolding super-resolution network (USRNet) \cite{DBLP:conf/cvpr/ZhangGT20} also adopts the MAP framework but is based on the original degradation model in Equation (\ref{equ:kernel_model}), and the corresponding two sub-problems become super-resolving an LR image blurred by kernel $\boldsymbol{k}$ and denoising an HR image with a virtual noise level \emph{$\mu$}. It enhances the solution framework by unfolding the iterative optimization process of DPSR into an end-to-end trainable network with iterative scheme, enabling joint optimization between the two sub-problems. A comparison between solution frameworks of DPSR and USRNet is depicted by Fig.\ref{fig:3_1_1_struct}(b) and (c). 
Besides, some other methods exploiting plug-and-play technique include \cite{DBLP:conf/icip/BrifmanRE16, DBLP:journals/tci/ChanWE17, DBLP:conf/cvpr/ZhangZGZ17}.

\vspace{8pt}
\textbf{Limitation}: In spite of the aforementioned progress, this kind of methods have one obvious drawback: they all rely on an additional input of degradation estimation, especially the SR kernel $\boldsymbol{k}$. However, estimating the correct kernel from an arbitrary LR image is not an easy task, and an inaccurate estimation input will cause kernel mismatch and greatly undermine the SR performance \cite{gu2019blind,shocher2018zero}. Fig.\ref{fig:limit_3_1_1} shows the comparison between SR results with correct and incorrect kernels based on the method SRMD. 
Therefore, only if one has some method at hand for reliable degradation estimation can he quickly obtain a satisfactory SR output, otherwise he may come to the tedious work of manually choosing a proper estimation input for better result.
Hence, we introduce another kind of approaches in the next part, which incorporate kernel estimation into SR framework for more robust performance.

%%-------------------------------------------------------------------------
\subsubsection{Image-Specific Adaptation with Kernel Estimation}\label{sec:one_one_two}

Iterative kernel correction (IKC) \cite{gu2019blind} proposes to correct kernel estimation in an iterative way to gradually approach a satisfactory result. The highlight of this method is to take advantage of the intermediate SR results, since the artifacts within an SR image caused by kernel mismatch tend to have regular patterns. Specifically, a corrector network is used to estimate the kernel correcting residual given an SR image conditioned on the current kernel. Then the updated kernel is used to generate a new SR result with fewer artifacts. The SR network includes spatial feature transform \cite{wang2018recovering} layers in each residual block, and the current kernel is used to generate transforming parameters for feature adaptation, which can be more effective than direct concatenation of inputs as proposed by SRMD. In addition, a predictor network is applied for kernel initialization based on the input LR image alone, and dimensionality stretching is adopted for kernel coding.
A more recent work, deep alternating network (DAN) \cite{huang2020unfolding}, further enhances the IKC framework. It unifies the corrector and SR network into an end-to-end trainable one instead of training each sub-network separately as IKC does. This joint training strategy can make the two networks more compatible to each other. Moreover, the corrector uses original LR input for kernel estimation conditioned on an intermediate SR result, which is beneficial to more robust kernel estimation performance.
The overall frameworks of IKC and DAN are illustrated in Fig.\ref{fig:3_1_2_struct}(a) and (b).
The idea of making use of SR artifacts for kernel estimation is also employed in variant blind SR (VBSR) \cite{cornillere2019blind}, yet it trains a kernel discriminator to estimate the error map of an SR output instead of the kernel itself, and finds the optimal kernel by minimizing the error of SR output during the inference stage, as shown by Fig.\ref{fig:3_1_2_struct}(c).
In addition to the SR kernel, estimation on more degradation types has also been studied. CBSR \cite{liu2020learning} combines two sub-networks for noise and kernel estimation with a non-blind SR network, thus forming a unified cascaded architecture for blind SR. 

In fact, the iterative scheme adopted by IKC and DAN can be interpreted well from the perspective of domain adaptation: instead of producing the final SR output in a single stroke like SRMD, it chooses several intermediate SR results as interchange stations during the long trip from input LR to the target \emph{Natural HR} domain, passing across the domain gap in Fig.\ref{fig:domain_pic} step-by-step. These two methods can have more robust performance than SRMD framework depending on the accuracy of kernel estimation input.

Nevertheless, such an iterative scheme usually consumes more inference time and requires human intervention to choose the optimal number of iterations. To tackle these issues, some recent works propose non-iterative frameworks by introducing more accurate degradation estimation or more efficient feature adaptation strategies. Unsupervised degradation representation learning for blind SR (DRL-DASR) \cite{dasrudrl} tries to estimate the degradation information with a trainable encoder in the latent feature space, and the degradation encoder is trained with contrastive learning in an unsupervised manner. Specifically, LR samples with the same degradation as the query input are considered as the positive examplars while those with different degradations are taken as negative ones. Then the mutual information among all samples is maximized in the latent space, leading to content-invariant degradation representations. Moreover, the estimated degradation representation is used to generate the corresponding convolutional kernels and modulation coefficients in SR network. Such a framework can achieve satisfactory SR results with a single forward pass. Kernel-oriented adaptive local adjustment (KOALAnet) \cite{koalanet} also utilizes a similar dynamic kernel strategy that adapts the SR network to a specific degradation, and it further extends the non-iterative framework to spatially-variant degradation with a downsampling network for local kernel estimation. 
Another work, adaptive modulation network with reinforcement learning (AMNet-RL) \cite{amnet_RL}, proposes a modified version of adaptive instance norm (AdaIN) \cite{huang2017adain} to incorporate kernel estimation into the SR network, and it also pioneered in optimizing the blind SR model with in-differentiable perceptual metrics (e.g., NIQE \cite{mittal2012niqe}) under reinforcement learning framework.

%\paragraph{}
There are also some other approaches proposing to learn a blind SR model by merely covering more degradations in the training dataset, especially more realistic kernels estimated from real images.
For instance, kernel modelling super-resolution (KMSR) \cite{zhou2019kernel} builds a large kernel pool with data distribution learning based on some realistic SR kernels estimated from real LR images. Kernels from this pool are then used to synthesize HR-LR training pairs according to the classical degradation model, and the training process just follows non-blind setting with supervised learning.
Usually, a more general training dataset enables the SR model to implicitly distinguish and adaptively deal with LR inputs with different degradations. 
In other words, the SR model will be implicitly endowed with more capacity for kernel estimation in the training process, thus avoiding explicit kernel estimation in the framework. However, such a direct way may not lead to top performance, as have been argued in \cite{zhang2018learning}.
A similar strategy is employed in RealSR \cite{Ji_2020_CVPR_Workshops} and RealSRGAN \cite{ren2020real} to build more generic training dataset with more kinds of realistic kernels. This process is presented in Fig.\ref{fig:3_1_2_struct}(d).
Besides these methods, a correction filter \cite{hussein2020correction} is designed for modifying an LR input to match the SR model with a pre-defined degradation, which is primarily based on the kernel estimation from LR.

\vspace{8pt}
\textbf{Limitation}: Compared with approaches without kernel estimation, these methods practically save us from efforts in searching for kernel estimation algorithms, especially during inference stage, and have demonstrated impressive performance. Yet, they still cannot avoid the inherent disadvantage of explicit modelling: they cannot give out satisfactory results for images with degradations not covered in their model. 
As presented in Fig.\ref{fig:limit_3_1_2}, for an SR model focusing on the degradation caused by kernel $\boldsymbol{k}$, like IKC, it can hardly deal with LR inputs with degradations out of its modelling scope. 
This limitation is really too tough for complex real-world images. Even if we are willing to retrain the model with more degradation types, it is impractical for us to explicitly model the degradation in an arbitrary LR and gather enough external training data, as stated in Sec.\ref{sec:realworld_image_type}.
Next, let us step into another type of methods, which utilize a single input image alone for image-specific SR modelling.

%%-------------------------------------------------------------------------

\subsection{Single Image Modelling with Internal Statistics}\label{sec:one_two}

\begin{figure*}[htbp]
\renewcommand{\captionfont}{\small \rmfamily}
\begin{center}
\includegraphics[width=0.98\linewidth]{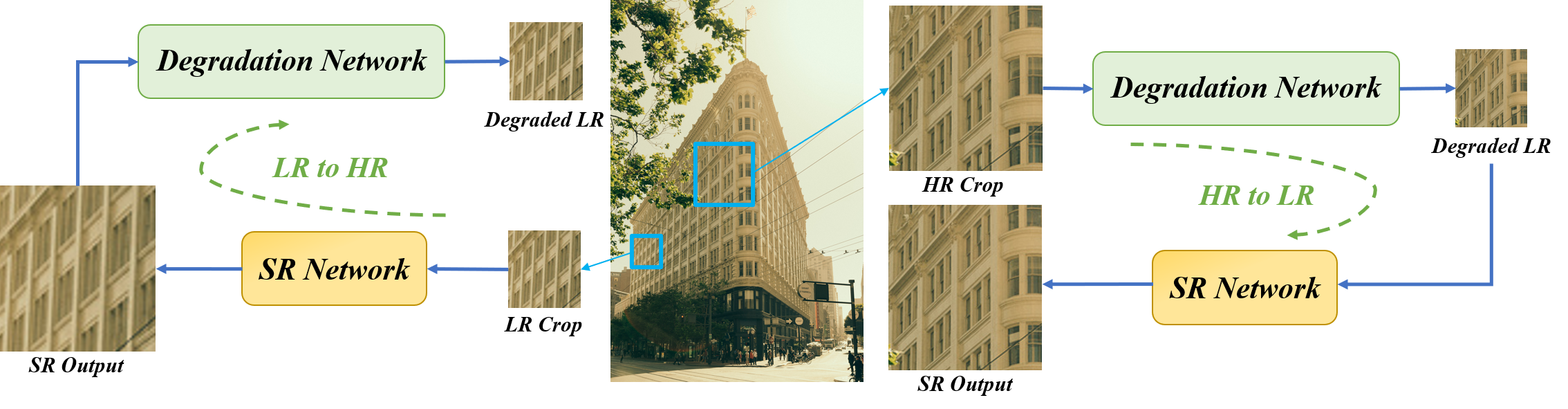}
\end{center}
\vspace{-10pt}
\caption{Network structure of DGDML-SR \cite{chengzero}. Two training paths are included, which are based on HR and LR crops from different image depths, respectively.}
\label{fig:3_2_struct}
\end{figure*}

SR modelling with a single image is based on the internal statistics of natural images: patches of a single image tend to recur within and across different scales of this image. The internal statistics has been quantified and proved to have more predictive power than external statistics for many natural images \cite{zontak2011internal}. A theoretical formulation is given by \cite{Michaeli_2013_ICCV}.
Specifically, one may assume that an HR image $\boldsymbol{h}$ and its LR counterpart $\boldsymbol{l}$ are taken by the same camera, but with an \emph{s}-scale zoom-in for the latter:
\begin{equation}
\label{equ:npbsr_model_1}
\centering
\boldsymbol{h}[n]=\int I(x)b_H(\frac{n}{s} -x)dx,
\end{equation} 
\begin{equation}
\label{equ:npbsr_model_2}
\centering
\boldsymbol{l}[n]=\int I(x)b_L(n-x)dx,
\end{equation} 
where \emph{I}(\emph{x}) is the continuous-space image, \emph{b} is the point spread function (PSF) of the camera, and \emph{b$_{H}$} should be a downscaled version of \emph{b$_{L}$} in the case of optical zoom:
\begin{equation}
\label{equ:npbsr_model_3}
\centering
b_H(x)=s b_L(s x).
\end{equation} 
Also, using the classical degradation modelling in Equation (\ref{equ:kernel_model}) without noise $\boldsymbol{n}$, the relationship between $\boldsymbol{h}$ and $\boldsymbol{l}$ expressed in discrete form is:
\begin{equation}
\label{equ:npbsr_model_4}
\centering
\boldsymbol{l}[n]=\sum_{m}^{}\boldsymbol{h}[m]\boldsymbol{k}[s n-m].
\end{equation} 
Now, for a given LR image, let \emph{q} and \emph{r} be two local patches with recurring pattern \emph{P}(\emph{x}) in the continuous scene, where \emph{r} is \emph{s}-times larger than \emph{q}. Then there will be:
\begin{equation}
\label{equ:npbsr_model_5}
\centering
\boldsymbol{r}[n]=\int P(\frac{x}{s})b_L(n-x)dx=\int sP(x)b_L(n-sx)dx,
\end{equation} 
\begin{equation}
\label{equ:npbsr_model_6}
\centering
\boldsymbol{q}[n]=\int P(x)b_L(n-x)dx,
\end{equation}
based on Equation (\ref{equ:npbsr_model_3}), one can finally arrive:
\begin{equation}
\label{equ:npbsr_model_7}
\centering
\boldsymbol{q}[n]=\sum_{m}^{}\boldsymbol{r}[m]\boldsymbol{k}[s n-m],
\end{equation}
which means that the relationship between \emph{q} and \emph{r} in the same LR is equivalent to two patches from an HR and its LR version related by kernel $\boldsymbol{k}$.
This property can be used to estimate $\boldsymbol{k}$ and solve the unknown HR.

Glasner et al. \cite{DBLP:conf/iccv/GlasnerBI09} did the pioneering work in 2009 to introduce internal statistics into solving SISR problem from a single image. Latter, nonparametric blind SR (NPBSR) \cite{Michaeli_2013_ICCV} further extends this framework to blind SR setting. Specifically, it proposes an MAP framework to estimate the SR blur kernel, based on the observation from Equation (\ref{equ:npbsr_model_7}) that the optimal kernel $\boldsymbol{k}$ is the one that maximizes the similarity among recurring patches across different scales.
In addition, NPBSR proves that the optimal $\boldsymbol{k}$ is not PSF of the camera but should be one with a smaller width, contrary to the common sense in its era.

The recent development of GAN gives birth to a new realization of using patch recurrence for blind kernel estimation. KernelGAN \cite{bell2019blind} interprets the maximization of patch recurrence within a single image as a data distribution learning problem. It assumes that the downsampled version of an LR image generated by the optimal $\boldsymbol{k}$ should share the same patch distribution with the original LR. Under GAN framework, a deep linear network is used as generator to parameterize the underlying SR kernel, and a discriminator distinguishes generated patches from those in original LR image.  
Once the training finishes, one can explicitly obtain the kernel estimation by convolving together all convolutional filters in generator. It is worth noting that the training process relies merely on the input LR without any external dataset, which can be seen as self-supervised learning.
Flow-based kernel prior (FKP) \cite{flow_kernel_prior} develops a more effective approach for kernel optimization, where a kernel prior in latent space is learned with normalizing flow (NF) \cite{DBLP:journals/corr/DinhKB14, DBLP:conf/iclr/DinhSB17} technique. Thanks to the invertible mapping between latent and pixel spaces enabled by NF, the search for the optimal $\boldsymbol{k}$ can be conducted in the learned kernel manifold. This process can be more efficient than directly optimizing a randomly initialized deep network, thus leading to more robust kernel estimation results.

The idea of self-supervision based on patch recurrence property can also be directly applied to performing SR. Zero-shot super-resolution (ZSSR) \cite{shocher2018zero}, developed by authors from the same group as NPBSR and KernelGAN, made the very first attempt to train an image-specific CNN for super-resolving each input LR without any pre-training step. The training is conducted with HR-LR pairs generated from a single LR input $\boldsymbol{y}$, where $\boldsymbol{y}$ is regarded as HR and coarser LR images are produced by downsampling with kernel $\boldsymbol{k}$. Data augmentation is utilized to make full use of information from the input image alone. The CNN trained with these image pairs will be capable of inferring specific relationships across different scales of $\boldsymbol{y}$, which is then used to super-resolve $\boldsymbol{y}$. In addition, ZSSR can be more robust to distracting artifacts (e.g., Gaussian noises, JPEG artifacts) by adding some noise to LR training samples, since it argues that only correlated image contents tend to recur across scales rather than noises.

In fact, ZSSR is still not well designed for blind setting: it requires estimated SR blur kernel $\boldsymbol{k}$ as input to guide the generation of coarser LR images for training. A unified self-supervision framework is thus proposed in DGDML-SR \cite{chengzero} - depth guided degradation model for learning-based SR. It combines a degradation network and an SR network into a single architecture, where the former is trained to simulate the degradation process, similar to the function of KernelGAN, and the latter aims to perform SR task like ZSSR does. This joint framework allows directly using generated LR as input to SR network without explicit extraction of SR kernel. In addition, DGDML-SR proposes to sample HR and LR patches in an unpaired way according to the depth map of input image, assuming that patches with smaller depth is equivalent to HR and those with larger depth to LR. A two-cycle training scheme similar to CycleGAN \cite{DBLP:conf/iccv/ZhuPIE17} structure is employed to simultaneously train the two networks (see Fig.\ref{fig:3_2_struct}), where the unpaired HR and LR patches are used as real samples for data distribution learning.

\vspace{8pt}
\textbf{Limitation}: The idea of self-supervision with internal statistics seems attractive for solving SR images from LR with variant degradation types, since it requires no effort in gathering large external training dataset. Nevertheless, its basic assumption may easily fail, especially for natural images with diverse contents (e.g., animals) or monotonous scenes (e.g., sky), since it is hard to exploit recurring information across scales to robustly perform SR with this kind of input images. Hence, these approaches can only produce favourable SR outputs for a very limited set of images with frequently recurring contents across scales, and new methods for single-image modelling are waiting to be explored for more general natural images.

\begin{figure*}[htbp]
\renewcommand{\captionfont}{\small \rmfamily}
\begin{center}
\includegraphics[width=1.01\linewidth]{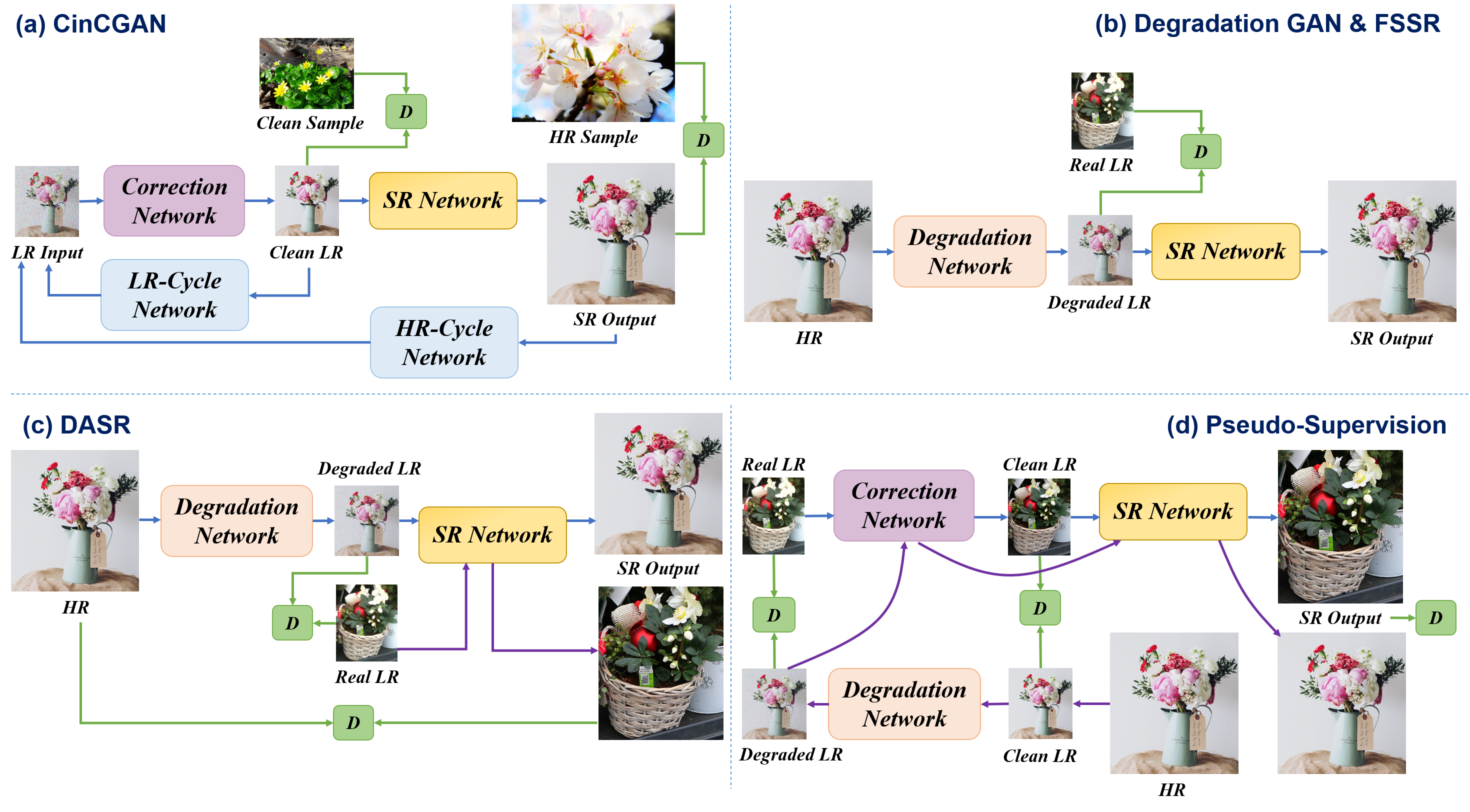}
\end{center}
\vspace{-10pt}
\caption{Overall architectures of methods with implicit modelling for data distribution learning. (a)CinCGAN \cite{yuan2018unsupervised}; (b)Degradation GAN \cite{Bulat_2018_ECCV} and FSSR \cite{fritsche2019frequency}, FS-SRGAN \cite{zhou2020guided}; (c)DASR \cite{dasr}; (d)pseudo-supervision \cite{maeda2020unpaired}. "D" represents discriminator network with GAN framework.}
\label{fig:GAN_domain}
\end{figure*}

\begin{figure*}[htbp]
\renewcommand{\captionfont}{\small \rmfamily}
\begin{center}
\includegraphics[width=1.0\linewidth]{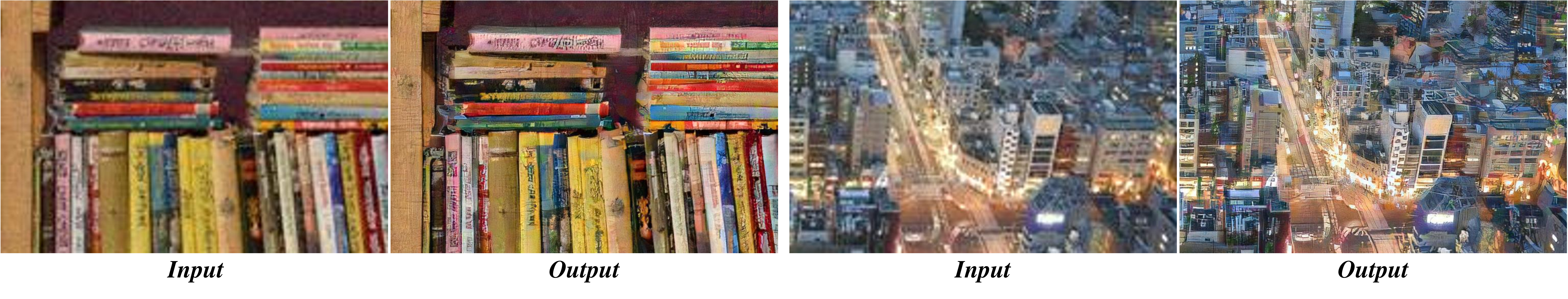}
\end{center}
\vspace{-12pt}
\caption{Illustration on the limitations of \emph{\textbf{Learning Data Distribution within External Dataset}}. For GAN-based framework, the SR results usually include severe artifacts and fake textures. Best viewed on screen.}
\label{fig:limit_4_1}
\end{figure*}

So far, we have had an overview on approaches with explicit degradation modelling, as well as their merits and demerits. Explicit modelling of degradation process is clear and straightforward, yet it can be too simple to model more complex degradations other than blurring and additive noise, such as real-world degradations originating from camera sensors. In fact, real-world images usually include multiple degradations, and we can hardly express these entangled factors with an explicit well-defined function. Hence, another group of methods propose to implicitly model the degradations through data distribution learning. To the best of our knowledge, so far there have only been approaches based on external dataset for implicit modelling, and we will talk about them in the following section.

%%-------------------------------------------------------------------------
\section{Implicit Degradation Modelling}\label{sec:two}

\subsection{Learning Data Distribution within External Dataset}

This kind of approaches aim to implicitly grasp the underlying degradation model through learning with external dataset. For dataset with paired HR-LR images, supervised learning with cautious design of SR network may be already enough to achieve satisfactory results, just like the top solutions proposed in NTIRE 2018 \cite{Timofte_2018_CVPR_Workshops} and AIM 2020 \cite{DBLP:conf/eccv/WeiLTLZPLXFZLHD20} challenges.
A more difficult setting is learning with unpaired data, where ground truth of LR images with realistic degradations are unavailable. Existing approaches usually exploit data distribution learning with GAN framework \cite{gan}, and one or more discriminators are used to distinguish generated image samples from real ones, pushing the generator towards the appropriate modelling direction. In most cases, two datasets are used to train the model, including HR and unpaired LR respectively, and one may regard the two datasets as representing target and source domains for learning the domain adaptation.

Among the earliest attempts on implicit modelling with unpaired data is CinCGAN \cite{yuan2018unsupervised}. It proposes to first transform an LR input to the \emph{Bicubic LR} domain before performing SR with a pre-trained non-blind model. The corresponding adaptation process is illustrated in Fig.\ref{fig:GAN_domain}(a). The \emph{Bicubic LR} domain is also regarded as \emph{Clean LR}, because its samples are generated with bicubic downsampling from HR images and assumed to be without any noise. Two CycleGAN structures \cite{DBLP:conf/iccv/ZhuPIE17} are respectively applied to transformation from \emph{LR} to \emph{Clean LR} and to target \emph{HR}, helping to maintain cycle consistency in the transformation process. In this way no paired data is required during training, thus forming an unsupervised training scheme. However, unsupervised domain adaptation is not an easy task, in that it is usually hard for a single 0/1 plane modelled by a discriminator to separate the right target domain.

To leverage both supervised learning and data distribution learning with GAN, some approaches focus on learning the degradation process from HR to LR, and the generated LR samples with realistic degradations are used to train the SR model in a paired manner. This process is depicted in Fig.\ref{fig:GAN_domain}(b). Degradation GAN \cite{Bulat_2018_ECCV} adopts this scheme and combines a High-to-Low degradation generator with a Low-to-High SR one in a single framework. Specifically, the High-to-Low generator simulates the degradation from \emph{HR} to \emph{LR} domain with an \emph{LR} discriminator, and the degraded LR image is used as input to Low-to-High generator for SR training. Adversarial loss is designed to dominate the training process, different from SRGAN \cite{DBLP:conf/cvpr/LedigTHCCAATTWS17} and ESRGAN \cite{esrgan} which mainly use pixel-wise loss for supervised training. Such learning strategy - unpaired degradation and paired SR - can also be found in some later works, including frequency separation for real-world SR (FSSR) \cite{fritsche2019frequency} and frequency separation SRGAN (FS-SRGAN) \cite{zhou2020guided}. These two methods both apply adversarial loss only to high-frequency contents in order to relieve the difficulty in adversarial training. FS-SRGAN further introduces a color attention module into its High-to-Low generator to alleviate color shifting problem during domain adaptation.

However, these approaches still cannot avoid the problems related to GAN framework. The generated LR images may have a large domain gap from real LR samples, thus undermining SR training performance. Towards addressing this issue, DASR \cite{dasr} proposes a domain-gap aware training strategy, where both generated LR images and real LR samples from dataset are used to train the SR model (see Fig.\ref{fig:GAN_domain}(c)). Another strategy called domain-distance weighted supervision is also employed to assign different loss weights to the LR inputs according to their domain distance to real \emph{LR}, helping to reduce negative influence caused by generated far-away LR samples. Specifically, predictions from \emph{LR} discriminator are used to quantitatively measure the domain distance for each LR.
Another work, pseudo-supervision \cite{maeda2020unpaired}, combines the forward SR reconstruction path in CinCGAN architecture with degradation learning to deal with problems caused by domain gap. This approach keeps the forward adaptation route of CinCGAN from degraded LR input to SR output, and adds a supervised path to it, as shown by Fig.\ref{fig:GAN_domain}(d). This supervised path starts from HR to degraded LR, then goes back to HR across the intermediate \emph{Clean LR} domain. This idea is in fact somewhat similar to the domain-gap aware training in DASR, if the CinCGAN route with \emph{Real LR} is regarded as the additional path of real LR images besides the supervised one.

\vspace{8pt}
\textbf{Limitation}: Though seemingly flexible and powerful, this kind of methods is still far from a cure-all in blind SR. On the one hand, these methods must rely on large external datasets to learn the SR model through implicit data distribution, but this data-hungry manner is not suitable for certain tasks, including old photo restoration. 
On the other hand, most of them exploit GAN framework for unsupervised data distribution learning. The GAN-based framework may be difficult to train, and it will frequently produce severe artifacts in SR results. These artifacts are harmful to many real-world applications, such as high-definition display and old photo/film restoration. Readers can refer to Fig.\ref{fig:limit_4_1} for illustration on the performance of a GAN-based method, FSSR.

%-------------------------------------------------------------------------
\subsection{Implicit Modelling with a Single Image: a Future Direction}\label{sec:two_two}

The idea of implicit modelling seems promising to deal with complex real-world degradations, as long as the source LR and target HR datasets are provided. However, there is still a long way to go since existing approaches mainly rely on GAN framework for data distribution learning, and the artifacts caused by GAN are harmful to many real-world applications. Besides exploring for more robust generative models, another direction is also worth noting, which has never been proposed so far: implicit modelling with a single LR image, as revealed in Fig.\ref{fig:classes}. 

As stated in previous sections, existing approaches all have their limitations, especially for the outliers of internal statistics with complex real-world degradations. Examples of such kind include surveillance video, old photos and films, as demonstrated in Sec.\ref{sec:realworld_image_type}.
These images are commonly seen in our daily life, and they pose great challenges to existing methods: they not only lack patch redundancy across scales to serve as hint for explicit modelling, but also cannot be covered with a few external datasets due to the complexity of unpredictable degradations.
We argue that this research gap can be possibly filled by methods based on implicit modelling with a single input image.
Up till now, there have been no related work in this field, but we believe this is a worthwhile direction for future research.

The main difficulty lying in this direction is the lack of effective SR prior, and one possible solution is to apply human intervention as additional information. For example, image restoration network with modulation \cite{he2019modulating} can be used to manipulate the SR outputs with a proper controlling coefficient, or to manually choose a clear image with similar contents to the LR input as an SR reference. The key point of these proposals is to increase the amount of useful information to make blind SR possible.

%-------------------------------------------------------------------------
\section{Datasets and Competitions}
\label{sec:dataset_competition}

\subsection{Datasets}
A large portion of methods covered in this paper, especially those with explicit degradation modelling and external dataset, require HR-LR image pairs for solving and evaluating SR models. However, due to the difficulty of obtaining real paired data, so far there have been only a few real-world datasets and most methods still synthesize LR inputs from HR images. Sec.\ref{sec:synthe_dataset} discusses the common ways of building synthetic dataset, and Sec.\ref{sec:realworld_dataset} gives an introduction on a few available real-world datasets.

\subsubsection{Synthetic Dataset}\label{sec:synthe_dataset}

For methods with explicit degradation modelling, the process of synthesizing degraded LR images from HR ground truth usually follows Equation (\ref{equ:kernel_model}), where a kernel $\boldsymbol{k}$ or noise with level $\boldsymbol{\sigma}$ is sequentially applied to the HR image together with the downscaling operation. 
For kernel $\boldsymbol{k}$, Gaussian kernels have been the most widely adopted kernel type. A typical practice is introduced by SRMD \cite{zhang2018learning}. An isotropic Gaussian kernel can be generated with a kernel width uniformly sampled from a pre-defined range, while an anisotropic one is characterized with a covariance matrix $\boldsymbol{\Sigma}$, where the rotation angle of its eigenvectors and the corresponding eigenvalues determine the kernel shape. 
As for noise $\boldsymbol{n}$, additive Gaussian noise is mostly used to simulate real-world noises, and the level $\boldsymbol{\sigma}$ can also be sampled from a specific range. Examples of Gaussian kernels and different noise levels are included in Fig.\ref{fig:real_lr_im}(a).
In addition, one can also enlarge the dataset with more realistic kernel or noise types, just as done by KMSR \cite{zhou2019kernel} and RealSR \cite{Ji_2020_CVPR_Workshops}.

The most popular HR datasets in the non-blind setting are also employed in blind SR. For example, DIV2K \cite{agustsson2017ntire} and Flickr2K \cite{DBLP:conf/cvpr/TimofteAG0ZLSKN17} are often used for training, while Set5 \cite{DBLP:conf/bmvc/BevilacquaRGA12}, Set14 \cite{zeyde2010single}, BSD100 \cite{huang2015single} and Urban100 \cite{DBLP:conf/cvpr/HuangSA15} are usually for testing. Specifically, a blind SR benchmark \emph{DIV2KRK} is proposed in KernelGAN \cite{bell2019blind}, where each image in DIV2K validation set is blurred and downsampled by a randomly generated anisotropic Gaussian kernel with some multiplicative noise to simulate more complex degradation.
Other synthetic datasets with unknown degradation, e.g. DIV2K wild \cite{Timofte_2018_CVPR_Workshops}, are also used by some methods with implicit modelling like CinCGAN \cite{yuan2018unsupervised}.

\subsubsection{Real-World Dataset}\label{sec:realworld_dataset}
Up till now, there have been several real-world datasets with paired HR-LR images, and these datasets are built with carefully designed techniques and advanced digital devices.
Representatives are City100 \cite{DBLP:conf/cvpr/ChenXTZW19}, DRealSR \cite{DBLP:conf/eccv/WeiXLZYZL20} and RealSR \cite{DBLP:conf/iccv/CaiZYC019}. Among them, DRealSR is the largest one with around 800 image pairs for each scale factor.
Usually, an HR image and its corresponding LR observations are captured by adjusting the focal lengths of imaging devices, then HR-LR pairs are accurately aligned with image registration and color rectification. 
Some other real-world datasets without HR ground truth have also been used as source domain images for SR network inputs, like DPED\cite{DBLP:conf/iccv/IgnatovKTVG17} in NTIRE 2020 Real-World Image SR challenge\cite{lugmayr2020ntire}.
Compared with synthetic data, these real-world datasets serve as an important benchmark for investigating blind SR in real setting. 
However, building such a dataset is time-consuming and expensive, and also cannot cover all scenarios due to complicated variations among different imaging systems.

\subsection{Competitions}
In order to gauge and promote the development of superior solutions, some competitions have been held in the field of SISR, including some tracks with unknown degradation or blind SR setting. Here, we make a summary on previous competitions related to blind SR in Table \ref{tab:challenge}, hoping to reveal some research trends from another perspective.

As the first competition related to blind setting, NTIRE 2017 challenge on SISR goes one step ahead from bicubic downscaling and assumes unknown blur and decimation operations to get LR images. More complex degradation types are then introduced by succeeding competitions, including real-world degradations arising from digital cameras. These competitions help to probe the state-of-the-art in this field. However, since paired training dataset is provided, their top solutions were largely focused on network structure enhancement as for non-blind SR. 
AIM 2019 challenge made the first attempt to address the real-world setting where paired training data is unavailable, hoping to stimulate research endeavours towards unsupervised learning. It can be seen that rising attention has been paid to this emerging task, especially from its follower NTIRE 2020, and the research community is expected to propose more novel solutions to fit into real-world problems.

\begin{table*}[htbp]
\renewcommand{\captionfont}{\small \rmfamily}
\normalsize
\caption{Details of challenges on blind image super-resolution. "T" denotes "track", "Synthe" means synthetic data generated from HR images, and --- indicates the same with previous row.}
\vspace{-8pt}
\label{tab:challenge}
\begin{center}
\begin{tabular}{p{95pt}p{145pt}p{45pt}cp{148pt}}
\hline
Competition Name                            & Task \& Degradation (Degrad.)                               & Train Set & Teams & Champion Solution \\[1.5pt] \hline \hline
\multirow{4}{*}{NTIRE 2017 SISR\cite{DBLP:conf/cvpr/TimofteAG0ZLSKN17}}                          & T2: Unknown downscaling, only blur        & DIV2K Synthe. (paired)        & \multirow{4}{*}{17}      & EDSR\cite{lim2017enhanced}: modified SRResNet + a downsampling network to learn the degradation for new training data generation with Flickr2K           \\ \hline

\multirow{8}{*}{NTIRE 2018 SISR\cite{Timofte_2018_CVPR_Workshops}}                       & T2: Realistic mild condition, motion blur \& Poisson noise, same degradation level for all images  & \multirow{3}{*}{---}        & \multirow{3}{*}{18}      & \multirow{5}{*}{\shortstack{WDSR\cite{DBLP:journals/corr/abs-1808-08718}: modified EDSR + pre-\\alignment to reduce random shift \\effects between HR-LR pairs}}        \\ \cline{2-4}
& T3: Realistic difficult condition, stronger degradation than T2  & \multirow{2}{*}{---} & \multirow{2}{*}{17}      &        \\  \cline{2-5}
& T4: Realistic wild condition, different degradation levels for each image  & ---        & \multirow{3}{*}{14}      & PDN\cite{pdn_ntire2018_track4}: PolyDense block motivated by PolyNet and DenseNet        \\ \hline

\multirow{4}{*}{\shortstack{NTIRE 2019 Real Im-\\age SR\cite{DBLP:conf/cvpr/CaiGTZa19}}}                  & Real-world SISR proposed in \cite{DBLP:conf/iccv/CaiZYC019}, real-world HR-LR image pairs captured with different focal lens of DSLR cameras                      & RealSR (paired)  & \multirow{4}{*}{36}    & UDSR\cite{udsr_ntire2019_track}: U-shaped network with U-Net structure + three-stage cascaded refinement framework with coarse to fine supervision       \\ \hline

\multirow{9}{*}{\shortstack{AIM 2019 Real-World \\Image SR\cite{DBLP:conf/iccvw/LugmayrJWKKHLHS19}}}   & T1: Same domain, synthetic degradation in source domain, SR results preserve low-level image characteristics, i.e. keep source domain quality               & Flickr2K Synthe., no target (unpaired)        & \multirow{5}{*}{7}      & \multirow{9}{*}{\shortstack{FSSR\cite{fritsche2019frequency}: DSGAN to learn to gen-\\erate LR images with degradations \\in source domain + ESRGAN train-\\ed with generated paired data + fre-\\quency-separation loss}}         \\ \cline{2-4}
& T2: Target domain, synthetic degradation in source domain, SR results should be of high-quality defined with target domain              & Flickr2K Synthe., DIV2K (unpaired)        & \multirow{4}{*}{4}      &          \\ \hline

\multirow{9}{*}{\shortstack{NTIRE 2020 Real Wo-\\rld Image SR\cite{lugmayr2020ntire}}}  & T1: Image processing artifacts, unknown synthetic degradation generated from image processing methods, source (input) and target domains are unpaired for training  & \multirow{5}{*}{---}        & \multirow{5}{*}{16}      & \multirow{9}{*}{\shortstack{RealSR \cite{Ji_2020_CVPR_Workshops}: build more generic pair-\\ed training dataset by covering more \\realistic kernels and noise types}}          \\ \cline{2-4}
& T2: Smartphone images, real images captured with a low-quality smartphone camera         & DPED\cite{DBLP:conf/iccv/IgnatovKTVG17} iPhone3, DIV2K (unpaired)        & \multirow{5}{*}{14}      &          \\ \hline
                                          
\multirow{4}{*}{\shortstack{AIM 2020 Real Image \\SR\cite{DBLP:conf/eccv/WeiLTLZPLXFZLHD20}}}                          & Real image SISR proposed in \cite{DBLP:conf/eccv/WeiXLZYZL20}, real natural images captured by five DSLR cameras with aligned HR-LR pairs         & DRealSR\cite{DBLP:conf/eccv/WeiXLZYZL20} (paired)        & \multirow{4}{*}{24}       & Proposed by Baidu team\cite{DBLP:conf/eccv/PanLXFZLHD20}: apply Neural Architecture Search among networks composed of dense residual blocks           \\ \hline
\end{tabular}
\vspace{-5pt}
\end{center}
\end{table*}

%-------------------------------------------------------------------------
\section{Quantitative Comparison}\label{sec:result}

This section includes detailed analysis on the performance and limitations of different methods with some testing examples. In fact, it is not an easy task to provide a comprehensive and fair comparison here, and we will state several problems hindering this practice in Sec.\ref{sec:benchmark_need}. Despite the difficulty, we still present some testing results based on officially released pre-trained models in Sec.\ref{sec:result_pretrained_model}, in order to provide readers with some useful information on the performance of each kind of methods. 

\subsection{Problem: Difficulty of Fair Comparison}\label{sec:benchmark_need}
\begin{itemize}
\setlength{\itemsep}{5pt}
\setlength{\parsep}{0pt}
\setlength{\parskip}{0pt}
\item \emph{\textbf{Inaccessible Code}}: Among the methods covered in our survey, some have released their full set of codes for training and testing, such as SRMD, IKC, RealSR, KernelGAN with ZSSR, and FSSR. 
But there are also some others which are not able to make their codes publicly available, and it is also non-trivial to reproduce these methods due to the lack of some important implementation details in original paper, especially for GAN-based approaches.

\item \emph{\textbf{Different Training Data}}: Even though the official implementations and pre-trained models are at hand, we still cannot sit back and relax for fair comparison since most of them adopt different datasets or different degradation types during training. These variables can greatly affect their performance and generalization capacity, especially for the blind setting with a large variety of distinct degradations. 
\end{itemize}

Hence, it is necessary and demanding to set up a benchmark for existing approaches. This benchmark should provide a fair and comprehensive comparison based on a uniform platform, and helps to gauge and push forward the state-of-the-art for blind SR. 
In this survey, we only show some quantitative results from published papers and a few testing examples produced by released pre-trained models.

\subsection{Comparison and Analysis Based on Pre-Trained Models}\label{sec:result_pretrained_model}
This section includes an overview on the quantitative performance of some representative approaches. Note that all the quantitative results shown in this section are copied from related papers. In order to be as fair as possible, every single table in this section is from a single paper, so results in the same table are ensured with a fair comparison. Also, each table mainly corresponds to a specific category in our proposed taxonomy. These results can provide readers with a straightforward comparison between different methods in each category.

For approaches with explicit modelling, we first show in Table \ref{tab:result_1} the comparison between two kernel estimation methods (NPBSR \cite{Michaeli_2013_ICCV} and KernelGAN \cite{bell2019blind} in Sec.\ref{sec:one_two}) combined with two SR models with kernel estimation input (SRMD \cite{zhang2018learning} in Sec.\ref{sec:one_one_one}, ZSSR \cite{shocher2018zero} in Sec.\ref{sec:one_two}). These results are reported in KernelGAN paper on the testing dataset DIV2KRK, which consists of 100 DIV2K validation images blurred and downsampled using randomly generated anisotropic Gaussian kernels. These results indicate that methods using degradation information as additional input can well be fitted into blind setting if combined with an appropriate kernel estimation algorithm. However, there still remains a considerable performance gap between using ground truth kernel and applying estimation methods, since it is non-trivial to accurately estimate degradation information from an arbitrary image. Also, these methods can hardly outperform those combining kernel estimation and SR into a single framework, such as DAN, showing the advantage of joint optimization.

The results of another two representative methods with explicit modelling and incorporated kernel estimation, namely IKC \cite{gu2019blind} and DAN \cite{huang2020unfolding} in Sec.\ref{sec:one_one_two}, are shown in Table \ref{tab:result_2}. These results are copied from paper of DAN \cite{huang2020unfolding}.
The testing data includes two classic datasets widely used in SR task, namely Set14 \cite{zeyde2010single} and BSD100 \cite{huang2015single}, and HR images are blurred and downsampled using eight selected isotropic Gaussian kernels for each scale factor to synthesize degraded LR. According to the comparison, DAN outperforms IKC as well as ZSSR for all the three scales.

\begin{figure*}[ht]
\renewcommand{\captionfont}{\small \rmfamily}
\begin{center}
\includegraphics[width=1.0\linewidth]{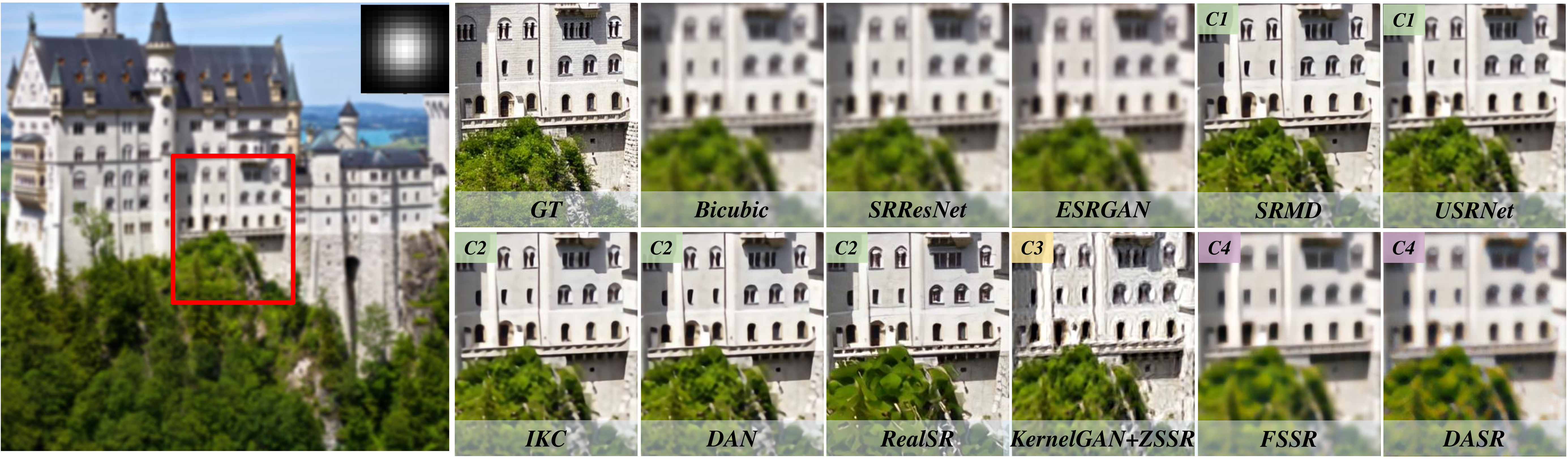}
\end{center}
\vspace{-16pt}
\caption{Qualitative comparison of synthetic testing example with Gaussian blur.}
\label{fig:result_pic_1}
\vspace{-8pt}
\end{figure*}

\begin{figure*}[ht]
\renewcommand{\captionfont}{\small \rmfamily}
\begin{center}
\includegraphics[width=1.0\linewidth]{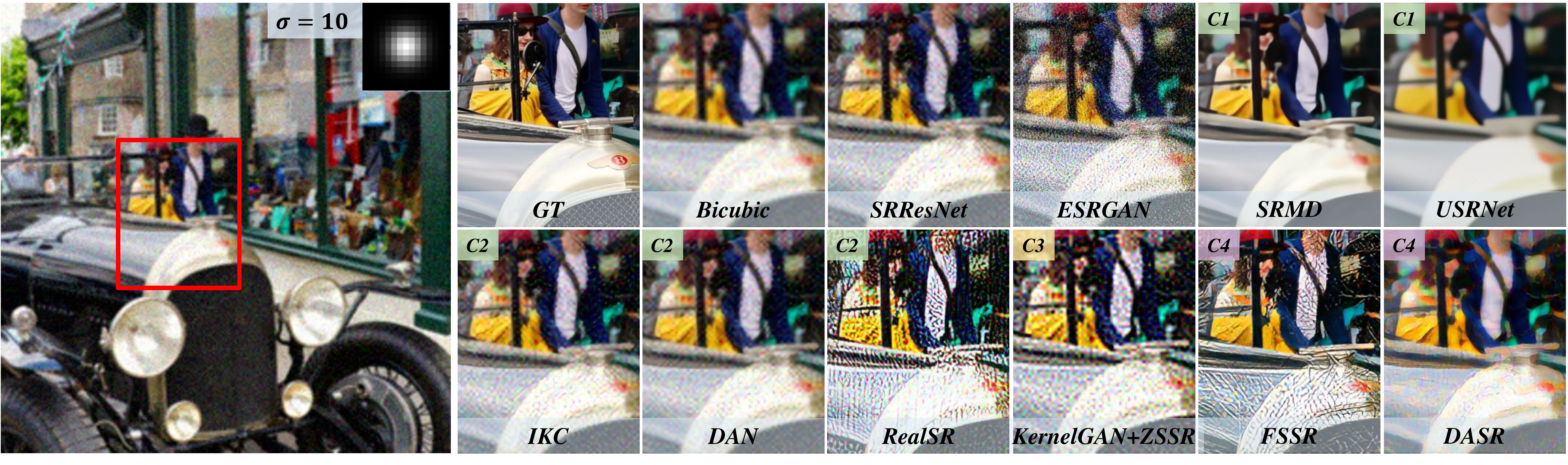}
\end{center}
\vspace{-16pt}
\caption{Qualitative comparison of synthetic testing example with Gaussian blur and noise.}
\label{fig:result_pic_2}
\vspace{-8pt}
\end{figure*}

\begin{figure*}[ht]
\renewcommand{\captionfont}{\small \rmfamily}
\begin{center}
\includegraphics[width=1.0\linewidth]{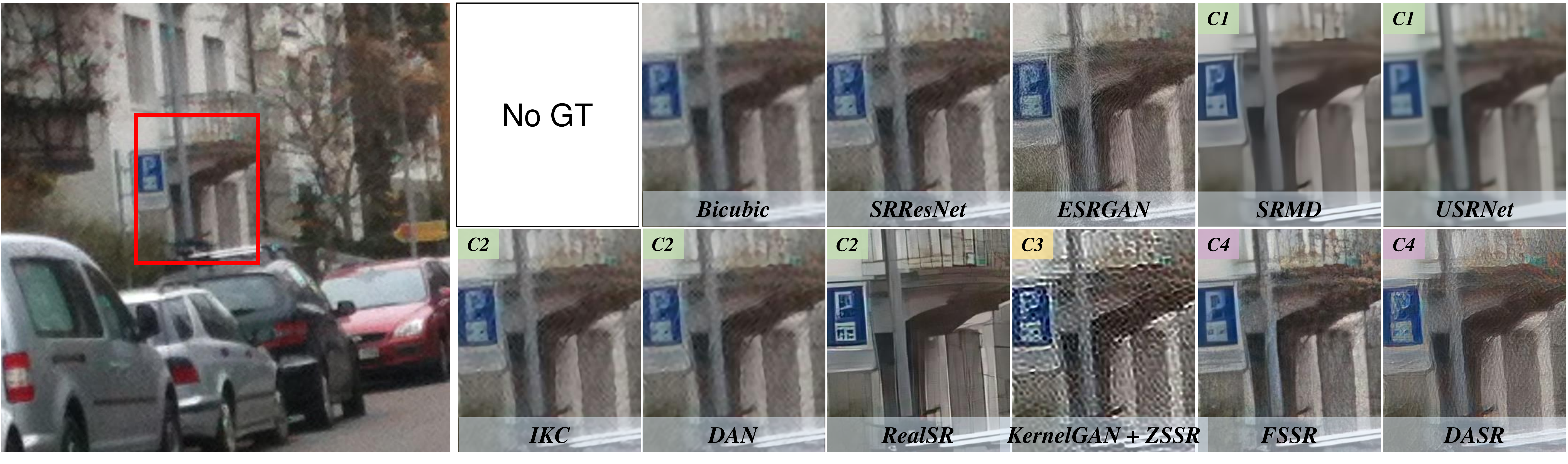}
\end{center}
\vspace{-16pt}
\caption{Qualitative comparison of real testing image captured with iPhone.}
\label{fig:result_pic_3}
\end{figure*}

For methods with implicit modelling, we show the results from DASR \cite{dasr} in Table \ref{tab:result_3}, including comparison among three representative data distribution learning models: CinCGAN \cite{yuan2018unsupervised}, FSSR \cite{fritsche2019frequency} and DASR \cite{dasr}. Besides PSNR and SSIM, LPIPS \cite{LPIPS} is used to better evaluate the perceptual quality of SR images generated from GAN. Two dataset are used for testing: AIM \cite{DBLP:conf/iccvw/LugmayrJWKKHLHS19} is released by AIM Challenge on Real World SR in ICCV 2019, while RealSR \cite{DBLP:conf/iccv/CaiZYC019} is a real-world dataset composed of HR-LR pairs captured with different focal lengths of the camera. We can see that DASR demonstrates the best performance, especially in terms of visual quality, owing to its better training strategies for narrowing the domain gap.

\begin{table}[htbp]
\renewcommand{\captionfont}{\small \rmfamily}
\normalsize
\caption{Quantitative comparison (PSNR(dB)/SSIM) of image-specific adaptation without kernel estimation (SRMD \cite{zhang2018learning}, ZSSR \cite{shocher2018zero}) plus blind kernel estimation (NPBSR \cite{Michaeli_2013_ICCV}, KernelGAN \cite{bell2019blind}).}
\label{tab:result_1}
\vspace{0pt}
\begin{center}
\begin{tabular}{lcll}
\hline
\multicolumn{1}{c}{Method} & \multicolumn{1}{c}{Scale}   & \multicolumn{1}{c}{DIV2KRK}                        \\ \hline \hline
Bicubic          & \multirow{9}{*}{2}             & 28.73 / 0.8040            \\ 
Bicubic kernel + \textbf{ZSSR}          &   & 29.10 / 0.8215          \\
\textbf{NPBSR} + \textbf{SRMD}      &   & 25.51 / 0.8083          \\
\textbf{NPBSR} + \textbf{ZSSR}      &   & 29.37 / 0.8370          \\
\textbf{KernelGAN} + \textbf{SRMD}                &   & 29.57 / 0.8564          \\
\textbf{KernelGAN} + \textbf{ZSSR}              &   & 30.36 / 0.8669          \\
Ground-truth kernel + \textbf{SRMD}     &   & 31.96 / 0.8955          \\
Ground-truth kernel + \textbf{ZSSR}     &   & 32.44 / 0.8992          \\ 
DAN   &  & 32.56 / 0.8997
\\
\hline
Bicubic          & \multirow{9}{*}{4}                 & 25.33 / 0.6795         \\ 
Bicubic kernel + \textbf{ZSSR}          &   & 25.61 / 0.6911         \\
\textbf{NPBSR} + \textbf{SRMD}      &   & 23.34 / 0.6530         \\
\textbf{NPBSR} + \textbf{ZSSR}      &   & 26.08 / 0.7138         \\
\textbf{KernelGAN} + \textbf{SRMD}               &   & 25.71 / 0.7265         \\
\textbf{KernelGAN} + \textbf{ZSSR}               &   & 26.81 / 0.7316         \\
Ground-truth kernel + \textbf{SRMD}     &   & 27.38 / 0.7655         \\
Ground-truth kernel + \textbf{ZSSR}     &   & 27.53 / 0.7446        \\
DAN   &  & 27.55 / 0.7582
\\ \hline
\end{tabular}
\vspace{0pt}
\end{center}
\end{table}

\begin{table}[ht]
\renewcommand{\captionfont}{\small \rmfamily}
\normalsize
\caption{Quantitative comparison (PSNR(dB)/SSIM) of image-specific adaptation with kernel estimation (IKC \cite{gu2019blind}, DAN \cite{huang2020unfolding}).}
\label{tab:result_2}
\vspace{0pt}
\begin{center}
\begin{tabular}{lclllll} \hline
\multicolumn{1}{c}{Method} & Scale      & \multicolumn{1}{c}{Set14} & \multicolumn{1}{c}{BSD100} \\ \hline \hline
Bicubic      & \multirow{4}{*}{2}     & 26.02 / 0.7634            & 25.92 / 0.7310                  \\
ZSSR                       &   & 28.35 / 0.7933            & 27.92 / 0.7632                      \\
\textbf{IKC}                        &   & 32.82 / 0.8999            & 31.36 / 0.9097                   \\
\textbf{DAN}                        &   & 33.07 / 0.9068            & 31.76 / 0.9213             
\\ \hline
Bicubic      & \multirow{4}{*}{3}    & 24.01 / 0.6662            & 24.25 / 0.6356                                    \\
ZSSR                       &    & 26.11 / 0.6942            & 26.06 / 0.6633                   \\
\textbf{IKC}                        &     & 29.46 / 0.8229            & 28.56 / 0.8493              \\
\textbf{DAN}                        &     & 30.09 / 0.8287            & 28.94 / 0.8606             \\ \hline
Bicubic      & \multirow{4}{*}{4}     & 22.79 / 0.6032            & 23.29 / 0.5786                         \\
ZSSR                       &      & 24.78 / 0.6268            & 24.97 / 0.5989                  \\
\textbf{IKC}                        &      & 28.26 / 0.7688            & 27.29 / 0.8014              \\
\textbf{DAN}                        &     & 28.43 / 0.7693            & 27.51 / 0.8078             \\ \hline                     
\end{tabular}
\vspace{0pt}
\end{center}
\end{table}

\setlength{\tabcolsep}{1.6mm}{}
\begin{table}[ht]
\renewcommand{\captionfont}{\small \rmfamily}
\normalsize
\caption{Quantitative comparison of implicit modelling with data distribution learning (CinCGAN \cite{yuan2018unsupervised}, FSSR \cite{fritsche2019frequency}, DASR \cite{dasr}). Note that ESRGAN \cite{esrgan} is trained with paired data. }
\label{tab:result_3}
\vspace{0pt}
\begin{center}
\begin{tabular}{lccc}
\hline
\multirow{2}{*}{Method} & \multicolumn{2}{c}{PSNR(dB) / SSIM / LPIPS$\downarrow$                                                } \\ \cline{2-3} 
                        & AIM                           & RealSR                            \\  \hline \hline
ESRGAN                  & - / - / -                  & 25.70 / 0.7487 / 0.20 \\
ZSSR                    & 22.33 / 0.6022 / 0.63 & 26.01 / 0.7482 / 0.39 \\
\textbf{CinCGAN}                 & 21.60 / 0.6129 / 0.46 & 25.09 / 0.7459 / 0.41              \\
\textbf{FSSR}                    & 20.82 / 0.5103 / 0.39 & 25.99 / 0.7388 / 0.27 \\
\textbf{DASR}                    & 21.60 / 0.5640 / 0.34 & 26.78 / 0.7822 / 0.23 \\  \hline
\end{tabular}
\vspace{0pt}
\end{center}
\end{table}

Qualitative comparison of some testing examples is shown in Fig.\ref{fig:result_pic_1}$\sim$\ref{fig:result_pic_3}, and we use pre-trained models provided by the authors of the corresponding papers. The first two degraded LR images are synthesized with isotropic Gaussian blur or additive Gaussian noise, and the third one is a real-world image from NTIRE real-world SR challenge \cite{lugmayr2020ntire} without ground truth. Note that for SRMD and USRNet, we directly use the ground truth kernel or noise level as additional inputs in order to validate the efficacy of the SR model alone. Also, for RealSR and FSSR, while there have been multiple pre-trained models based on different training datasets, we choose the one trained with real images taken by mobile devices from DPED dataset \cite{DBLP:conf/iccv/IgnatovKTVG17}. 
In addition, we include two renowned non-blind SR models, SRResNet \cite{ledig2017photo} and ESRGAN \cite{esrgan}, into the comparison list for readers' reference.
Based on these testing examples, some important observations can be drawn as following:

\begin{itemize}
\setlength{\itemsep}{5pt}
\setlength{\parsep}{0pt}
\setlength{\parskip}{0pt}
\item[(1)] For methods exploiting external dataset, their generalization largely depends on the coverage of degradation modelling or training data distribution. For example, approaches with explicit modelling can only handle noisy inputs if noise is directly covered as a degradation factor in the SR modelling, like SRMD and USRNet. Otherwise, they will not be capable of noise removal and consequently bring unfavourable artifacts in SR results, like SRResNet and IKC. On the other hand, models trained on real image dataset hardly give out visually pleasing results on synthetic data, such as RealSR and FSSR. 
\item[(2)] Real-world images \emph{do} include more complex degradations, which deviate a lot from synthetic data distribution. Models with explicit modelling generally perform well on the synthetic images within their degradation coverage, like SRMD and IKC. However, none of the methods generate satisfactory result for the third real image, including those with implicit modelling. Specifically, SRResNet and IKC tend to keep the noise texture and cause artifacts. SRMD and USRNet alleviate the noisy artifacts to some extent but in the sacrifice of high-frequency details, and it also needs some effort to estimate kernel and noise level for a real image with unknown degradations. RealSR and FSSR do have better results in terms of removing noise and preserving sharp textures, yet they still generate fake textures or artifacts, leading to the unnatural-looking of SR images.
\end{itemize}

\subsection{Suggestions on Fair Comparison}

We would like to suggest that our readers make better use of our proposed taxonomy in their future work, especially for effective and fair comparison among different methods in their paper. 
One may first try to place their own method into the corresponding category, and then pay special attention to previous methods belonging to the same category for evaluation and comparison. 
As for methods out of the specific categorial scope, since they have employed different degradation modelling or data sources (external or internal), direct comparison may become unfair and unnecessary, sometimes causing difficulty and even confusion.
Hence, we recommend here that future work may as well follow our taxonomy to make comparison among methods from the same category, both for proposing new solutions and benchmark setting.

%-------------------------------------------------------------------------
\section{Conclusion}\label{sec:conclusion}

In this paper, we present a systematic survey on recent progress in blind image SR. In order to effectively classify and summarize existing methods, we propose a taxonomy according to their ways of degradation modelling and the data used for solving the SR model: explicit or implicit modelling with external dataset or a single LR image. Except implicit modelling with a single image, the other three categories all have representative existing approaches, and we make a conclusion on them as following:

\begin{itemize}
\setlength{\itemsep}{5pt}
\setlength{\parsep}{0pt}
\setlength{\parskip}{0pt}
\item \emph{\textbf{Explicit modelling with external dataset}}: representatives are SRMD and IKC, which utilize the classical degradation model or its variants for image-specific adaptation based on degradation information. These methods perform well on degradation types covered in its modelling, but their performance will severely deteriorate on other degradations.
\item \emph{\textbf{Explicit modelling with a single LR image}}: including NPBSR and KernelGAN for blind kernel estimation, as well as ZSSR and DGDML-SR for SR. They leverage the internal statistics of natural images - patch recurrence across scales, which can also be theoretically derived from classical degradation model. However, these methods may fail for more general natural images with diverse or monotonous scenes, i.e., those without enough recurring clues for SR task.
\item \emph{\textbf{Implicit modelling with external dataset}}: such as CinCGAN, FSSR and DASR. These methods assume that real-world degradations are too complex to be explicitly modelled, but can be implicitly learned with data distribution learning under GAN framework. However, domain adaptation is a non-trivial task due to the large space of the natural image domain, making it hard to train a GAN-based model with good performance and generalization capacity.
\end{itemize}

From our perspective, implicit modelling with a single image, which has not been proposed yet, is a direction worth exploring in future research, especially for general natural images with complex degradations and without strong internal statistics. One possible solution to this problem is utilizing human intervention to provide additional information as SR prior, and restoration network with modulation or manually choosing an HR reference image may be of help. We hope this paper can inspire some new ideas for future work and make contributions to the prosperity of blind image SR techniques.

% or


%\appendices

\ifCLASSOPTIONcaptionsoff
  \newpage
\fi



%\bibliographystyle{IEEEtran}
%\begin{thebibliography}{1}

\bibliographystyle{IEEEtran}
\bibliography{egbib}

%\end{thebibliography}

\begin{IEEEbiography}
[{\includegraphics[width=1in,height=1.25in,clip,keepaspectratio]{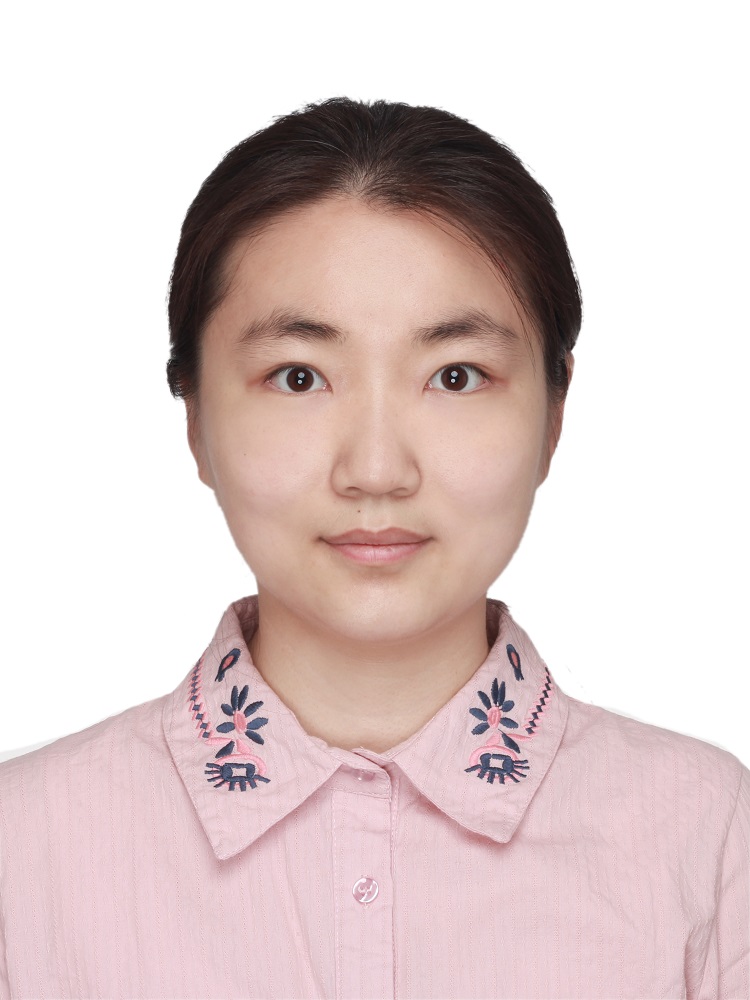}}]
{Anran Liu}
received the B.Eng. degree in 2019 from Tsinghua University, Beijing, China. She is now pursuing a Ph.D. degree at the Department of Computer Science, the University of Hong Kong. Her research interests include computer vision and deep learning, particularly focusing on image processing and super-resolution.
\end{IEEEbiography}

\begin{IEEEbiography}
[{\includegraphics[width=1in,height=1.25in,clip,keepaspectratio]{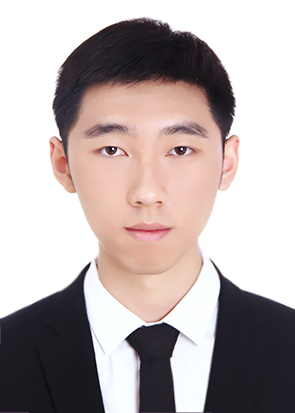}}]
{Yihao Liu}
received the B.S. degree from University of Chinese Academy of Sciences, Beijing, in 2018. He is now working towards the Ph.D. degree in Multimedia Laboratory, Shenzhen Institute of Advanced Technology, Chinese Academy of Sciences. He is supervised by Prof. Yu Qiao and Prof. Chao Dong. His research interests include computer vision and image/video enhancement.
\end{IEEEbiography}

\begin{IEEEbiography}
[{\includegraphics[width=1in,height=1.25in,clip,keepaspectratio]{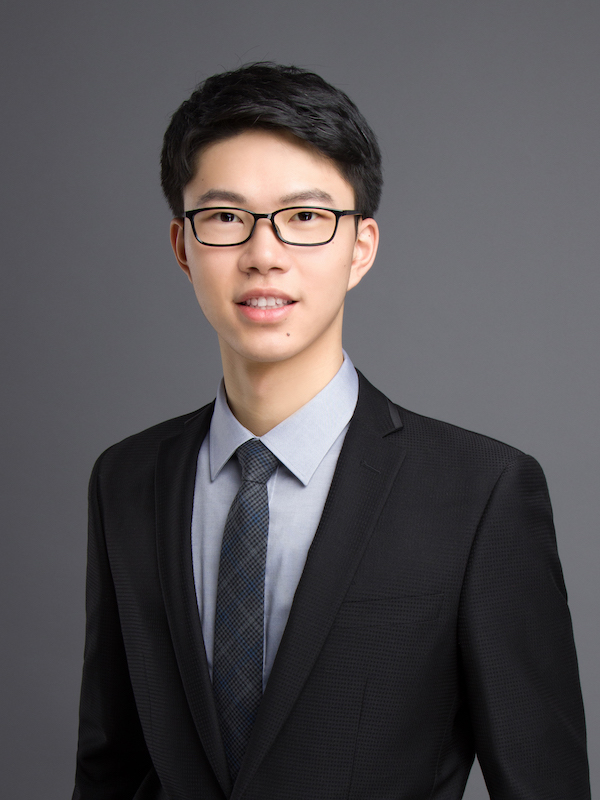}}]
{Jinjin Gu}
received his B.Eng. degree in computer science and engineering from the Chinese University of Hong Kong, Shenzhen, in 2020. He is currently pursuing a Ph.D. degree in Engineering and IT with the University of Sydney. His research interests include computer vision, image processing, interpretability of deep learning algorithms and the applications of machine learning in industrial.
\end{IEEEbiography}

\begin{IEEEbiography}
[{\includegraphics[width=1in,height=1.25in,clip,keepaspectratio]{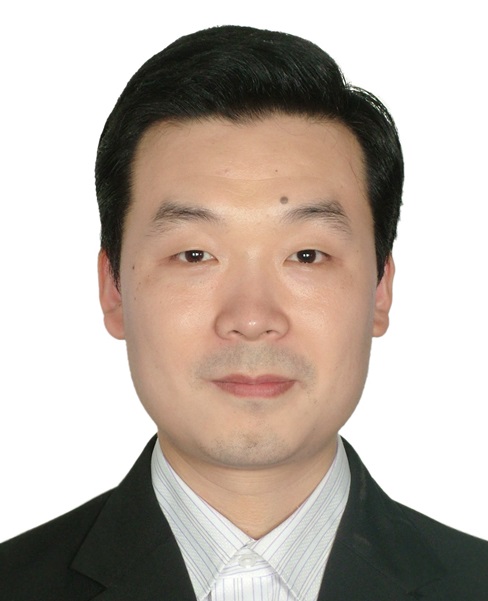}}]
{Yu Qiao}
(Senior Member, IEEE) is currently a Professor with the Shenzhen Institutes of Advanced Technology (SIAT), Chinese Academy of Science, and the Director of the Institute of Advanced Computing and Digital Engineering. He has published more than 180 articles in international journals and conferences, including T-PAMI, IJCV, T-IP, T-SP, CVPR, and ICCV. His research interests include computer vision, deep learning, and bioinformation. He received the First Prize of the Guangdong Technological Invention Award, and the Jiaxi Lv Young Researcher Award from the Chinese Academy of Sciences. His group achieved the first runner-up at the ImageNet Large Scale Visual Recognition Challenge 2015 in scene recognition, and the Winner at the ActivityNet Large Scale Activity Recognition Challenge 2016 in video classification. He has served as the Program Chair of the IEEE ICIST 2014.
\end{IEEEbiography}

\begin{IEEEbiography}
[{\includegraphics[width=1in,height=1.25in,clip,keepaspectratio]{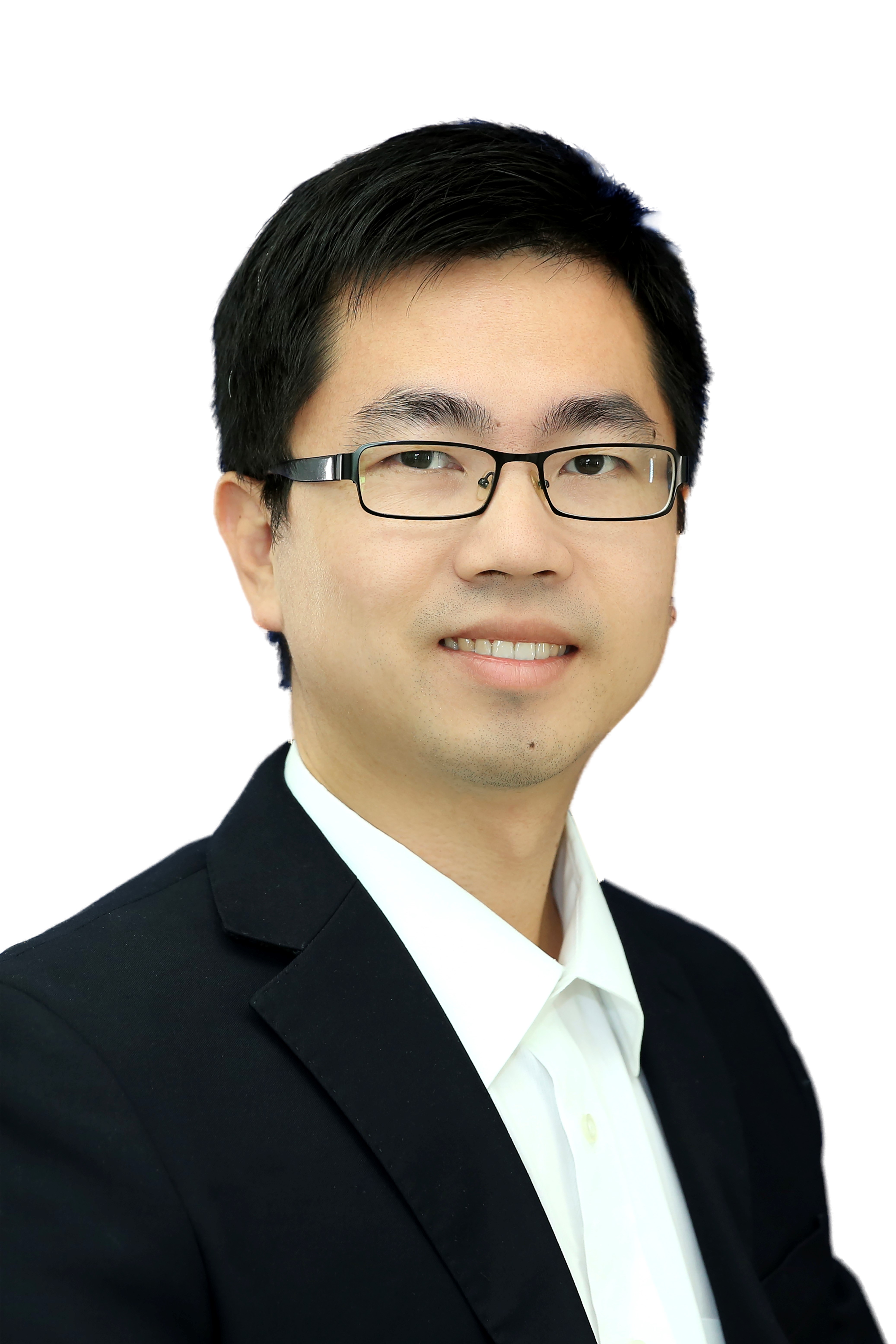}}]
{Chao Dong}
is currently an associate professor in Shenzhen Institute of Advanced Technology, Chinese Academy of Science. He received his Ph.D. degree from The Chinese University of Hong Kong in 2016. In 2014, he first introduced deep learning method -- SRCNN into the super-resolution field. This seminal work was chosen as one of the top ten “Most Popular Articles” of TPAMI in 2016. His team has won several championships in international challenges –NTIRE2018, PIRM2018, NTIRE2019, NTIRE2020 and AIM2020. He worked in SenseTime from 2016 to 2018, as the team leader of Super-Resolution Group. His current research interest focuses on low-level vision problems, such as image/video super-resolution, denoising and enhancement.
\end{IEEEbiography}

\end{document}